\documentclass{article}

\usepackage{microtype}
\usepackage{graphicx}
\usepackage{subcaption}
\usepackage{booktabs} 

\usepackage{hyperref}

\usepackage[preprint]{icml2026}

\usepackage{amsmath}
\usepackage{amssymb}
\usepackage{mathtools}
\usepackage{amsthm}

\usepackage[capitalize,noabbrev]{cleveref}

\theoremstyle{plain}

\theoremstyle{definition}

\theoremstyle{remark}

\usepackage[textsize=tiny]{todonotes}

\usepackage{graphicx}
\usepackage{algorithm}
\usepackage{algorithmic}
\usepackage{wrapfig}
\usepackage{enumitem}
\usepackage{booktabs}   
\usepackage{multirow}   
\usepackage{adjustbox}
\usepackage{siunitx}
\usepackage{caption} 
\usepackage{bm}
\usepackage{tikz}

\usepackage{xcolor}
\usepackage{colortbl} 
\usepackage{etoolbox} 
\usepackage{xcolor}
\usepackage{colortbl}
\usepackage{etoolbox}

\usepackage{xcolor}
\usepackage{colortbl}
\usepackage{etoolbox}

\newcommand{\tikzxmark}{%
\tikz[scale=0.23] {
    \draw[line width=0.7,line cap=round] (0,0) to [bend left=6] (1,1);
    \draw[line width=0.7,line cap=round] (0.2,0.95) to [bend right=3] (0.8,0.05);
}}

\begin{document}

\twocolumn[
  \icmltitle{Efficient Multi-Source Knowledge Transfer by Model Merging}



  \icmlsetsymbol{equal}{*}

\begin{icmlauthorlist}
    \icmlauthor{Marcin Osial}{uj,akces}
    \icmlauthor{Bartosz Wójcik}{uj}
    \icmlauthor{Bartosz Michał Zieliński}{uj}
    \icmlauthor{Sebastian Cygert}{nask,pg}
    \end{icmlauthorlist}
    
    \icmlaffiliation{uj}{Jagiellonian University, Kraków, Poland}
    \icmlaffiliation{nask}{NASK – National Research Institute, Warsaw, Poland}
    \icmlaffiliation{pg}{Gdańsk University of Technology, Gdańsk, Poland}
    \icmlaffiliation{akces}{Akces NCBR, Warsaw, Poland}
    
    \icmlcorrespondingauthor{Marcin Osial}{marcin.osial@uj.edu.pl}

  \icmlcorrespondingauthor{marcin.osial@doctoral.uj.edu.pl}

  \icmlkeywords{Machine Learning, ICML}

  \vskip 0.3in
]



\printAffiliationsAndNotice{}  

\begin{abstract}
While transfer learning is an effective strategy, it often overlooks the opportunity to leverage knowledge from numerous available models online. Addressing this multi-source transfer learning problem is a promising path to boost adaptability and cut re-training costs. However, existing methods remain inherently coarse-grained: they lack the precision needed for fine-grained knowledge extraction as well as the scalability required to aggregate knowledge from either large numbers of source models or models with high parameter counts. We address these limitations by leveraging Singular Value Decomposition (SVD) to first decompose each source model into its elementary, rank-one components. A subsequent aggregation stage then selects only the most salient components from all sources, thereby overcoming the previous efficiency and precision limitations. To best preserve and leverage the synthesized knowledge base, our method adapts to the target task by fine-tuning only the principal singular values of the merged matrix. In essence, this process recalibrates the importance of top SVD components. The proposed framework allows for efficient and scalable multi-source transfer learning in both vision and language domains, while remaining robust to perturbations in both the input space and the parameter space.
\end{abstract}


\section{Introduction}

The increasing complexity of models and the immense computational costs associated with their training necessitate the efficient utilization of existing resources. Transfer learning~\cite{zhuang2020comprehensive}, which involves initializing networks with weights from a pre-trained model, has emerged as a standard practice. This practice relies on foundational models, such as large-scale vision transformers~\cite{awais2025foundation} and self-supervised models~\cite{caron2021emerging}, which learn robust and generalized representations from vast, general-purpose datasets (e.g., ImageNet, LAION-5B). By effectively leveraging this broad pre-existing knowledge, transfer learning significantly reduces the demand for extensive task-specific data, accelerates downstream training, and enhances overall model performance across a wide range of computer vision tasks.

However, the wealth of specialized knowledge residing in other fine-tuned models remains largely untapped. Each model represents a valuable knowledge asset, with hundreds of thousands of versions publicly available on platforms like Hugging Face. 
Each new adaptation typically requires training from its original, pre-trained state, neglecting the specialized knowledge already acquired by previously fine-tuned models for distinct tasks. This gap has sparked considerable interest in developing methods for combining multiple models into a unified model~\cite{zootuning, yang2022deep}. Among these is model merging~\cite{yang2024model}, which presents an opportunity to fuse capabilities. 

A notable example is the aTLAS method~\cite{zhang2024knowledge}, which addresses the multi-source knowledge transfer to a new target task. It operates by learning to anisotropically combine task vectors~\cite{ilharco2022editing}, defined as the weight differences between fine-tuned models and their pre-trained states. The method operates by learning a distinct coefficient for each of the $T$ tasks, across each of the $L$ layers, and for each of the $P$ partitions within a weight matrix. These coefficients collectively form a learned tensor with dimensions $T \times L \times P$, allowing for adjustments to the model's behavior for new tasks. While holding significant promise, aTLAS lacks mechanisms for granular parameter selection, which restricts the precision of knowledge fusion.
Furthermore, aTLAS's memory footprint scales linearly with the number of added sources due to its reliance on using full task vectors. This design prevents the aggregation of larger models or a greater number of source models. As a result, its training is confined to multi-GPU environments, undermining its parameter-efficient benefits. This coarse-grained approach lacks a robust knowledge composition mechanism, making it susceptible to perturbations from both corrupted and pruned parameters and degraded inputs.
\begin{figure*}[t!]
    \centering
    \includegraphics[width=\linewidth]{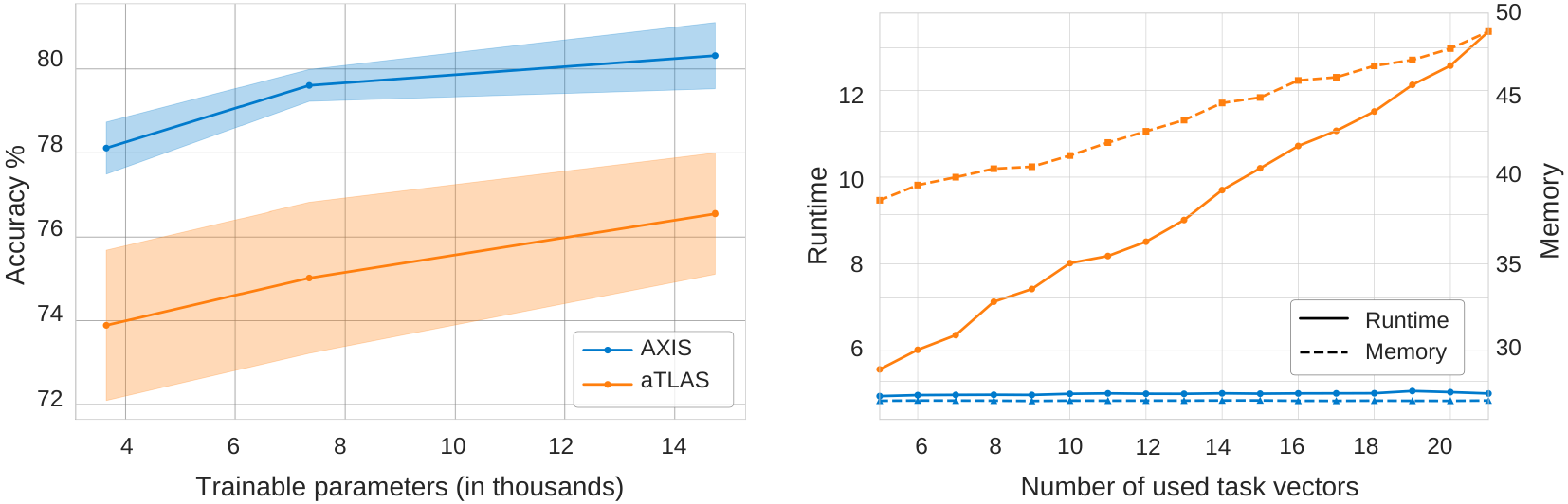}
    \caption{\textbf{Left}: Accuracy versus the number of trainable parameters (in thousands) for our method and aTLAS, averaged over all target tasks with ViT-B-32 architecture. Each data point corresponds to a fine-tuning parameter budget defined by the top N singular values (N=10\%, 20\%, and 40\%). The solid line denotes the mean accuracy, while the shaded area represents the standard deviation. The variation is calculated over all source task vectors.
    \textbf{Right}: Scalability analysis for ViT-L-14 architecture with $N$=10\% trainable parameters. As the number of source task vectors increases, the runtime and memory costs of aTLAS scale near-linearly. In contrast, our AXIS framework maintains a constant computational footprint.}
    \label{fig:param_efficiency}
\end{figure*}

In this paper, we propose a scalable approach that first aggregates knowledge and then allows for its efficient refinement during adaptation. First, we leverage Singular Value Decomposition (SVD) to decompose each task vector into its elementary, rank-one components. This allows us to identify and isolate granular patterns learned for each source task. Crucially, we identify these principal components as the primary carriers of transferable knowledge. A subsequent combination stage aggregates these components from all source models, performing a joint ranking to retain only a small, fixed number of the most significant ones. We term this strategy \textbf{AXIS}, as it embodies the principle of \textbf{A}ggregation by e\textbf{X}traction of \textbf{I}mportant \textbf{S}ingular components. Such selective aggregation ensures a stable memory usage and constant wall-time footprint during training, irrespective of the number of source models or original task matrix sizes (see Figure~\ref{fig:param_efficiency}). 
Consequently, our method yields intrinsic resilience to diverse and severe degradations, demonstrating superior stability compared to baselines by effectively mitigating the impact of compromised data signals, such as partial information or common corruptions, and challenging weight-space configurations, including pruning and noise injection serving as a proxy for corrupted source states. Our key contributions include:


\begin{itemize}[leftmargin=*]
\item We introduce a new approach to multi-source knowledge transfer, which outperforms the state-of-the-art method, aTLAS, across a wide spectrum of evaluation conditions, including 21 distinct tasks and various learning parameter budgets. Our evaluation encompasses three vision model scales and extends to the language domain.
\item We demonstrate AXIS's robustness to realistic and challenging parameter and input degradations.
\item The computational efficiency of AXIS is a key advantage, allowing for the scaling of knowledge transfer from a large number of source tasks and larger models.
\item Through ablation studies, we offer insights into the underlying structure of knowledge composition in AXIS and how it can be leveraged even for models with different pre-training history.
\end{itemize}

\begin{figure*}[t]
    \begin{minipage}[t]{0.50\textwidth}
        \vspace{0pt}
        \centering
        \includegraphics[width=0.99\columnwidth]{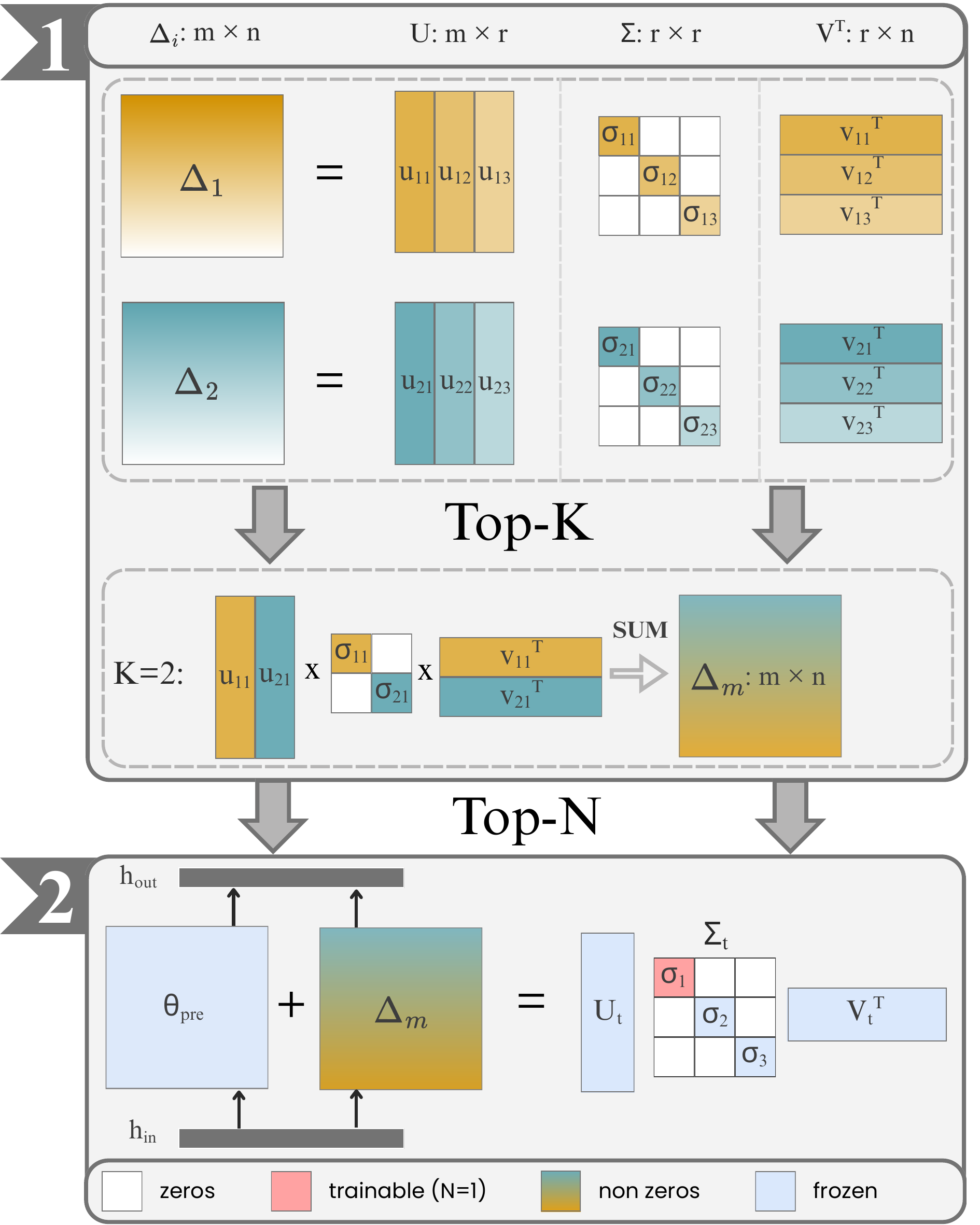}
    \end{minipage}
    \hfill
    \begin{minipage}[t]{0.46\textwidth}
        \vspace{0pt}
        \refstepcounter{algorithm}
        \textbf{Algorithm \thealgorithm} AXIS
        \label{alg:svd_agg} 
        \hrule
        \hrule
        \begin{algorithmic}[1]
    \STATE Initialize SVD components: $\mathcal{C} \leftarrow \emptyset$
    \FOR{each source task $i \in \{1, ..., T-1\}$}
        \STATE Compute the SVD of $\Delta_i = U_i \Sigma_i V_i^{\top}$
        \STATE $\mathcal{C} \leftarrow \mathcal{C} \cup \{ (\mathbf{u}_{j}, \sigma_{j}, \mathbf{v}_{j}^{\top}) \}_{j=1}^{r_i}$
    \ENDFOR 
    \STATE \textbf{Select} the top-K components to form $\mathcal{B}$
    \STATE \hspace*{1.5em} $\operatorname{Sort}_{\sigma_{k} \downarrow}(\mathcal{C}) \rightarrow \mathcal{B}$
    \STATE \textbf{Assemble} non-orthogonal vectors: 
    \STATE \hspace*{1.5em} $U_m \leftarrow [ u_1 | u_2 | \dots | u_K ]$
    \STATE \hspace*{1.5em} $\Sigma_m \leftarrow \operatorname{diag}(\sigma_1, \sigma_2, \dots, \sigma_K)$
    \STATE \hspace*{1.5em} $V_m \leftarrow [ v_1 | v_2 | \dots | v_K ]$
    \STATE \textbf{Reconstruct} from components:
    \STATE \hspace*{1.5em} $\Delta_m \leftarrow U_m \Sigma_m V_m^{\top}$
    \STATE \textbf{Re-orthogonalize} the basis via SVD:
    \STATE \hspace*{1.5em} $\Delta_m = U_{\text{t}} \Sigma_{\text{t}} V_{\text{t}}^{\top}$
    \STATE \textbf{Define} the set of learnable parameters $\Lambda$ as the top-$N$ singular values from $\Sigma_{\text{t}}$:
    \STATE \hspace*{1.5em} $\Lambda \leftarrow [s_1, \dots, s_N]$
    \STATE \textbf{Define} frozen singular values:
    \STATE \hspace*{1.5em} $\mathbf{s}_{\text{frozen}} \leftarrow \text{diag}(\Sigma_t) \setminus \Lambda$
    \STATE \textbf{Reconstruct} with learned values:
    \STATE \hspace*{1.5em} $\Delta_{\text{t}} \leftarrow U_{\text{t}} \operatorname{diag}(\Lambda, \mathbf{s}_{\text{frozen}}) V_{\text{t}}^{\top}$
    \STATE \textbf{return} $\Delta_{\text{t}}$   
    \end{algorithmic}
    \hrulefill
    \end{minipage}
    \caption{An overview of the AXIS framework. The process consists of two stages: 
    \textbf{(1) Extraction and aggregation:} Each source task matrix ($\Delta_1, \Delta_2, \dots$) is decomposed into its elementary singular components using SVD. The most salient components from all sources are selected based on a global top-$K$ ranking of their singular values. These K components are then summed to synthesize the merged task matrix, $\Delta_m$. For clarity, the diagram illustrates this with K = 2.
    \textbf{(2) Adaptation:} To form a stable and decorrelated transfer basis, $\Delta_m$ is re-parameterized via a final SVD. The model is then adapted to the target task by fine-tuning only a small subset (top-$N$) of the most principal singular values of the resulting matrix $\Sigma_t$ in each layer.
    }
    \label{fig:method_overview_phase_one}
\end{figure*}

\section{Related Work}

\textbf{Model merging} is gaining traction as a promising approach to leverage fine-tuned models without requiring access to training data or incurring an increase in model size and inference costs. The merging stage itself demands low computational resources and could be entirely training-free. While substantial progress has been made in combining models with diverse architectures~\cite{du2025adamms} or those trained without a shared initialization~\cite{rinaldi2025update, stoica2023zipit, ainsworth2022git}, our work primarily focuses on a distinct, yet highly prevalent paradigm where models originate from a common pre-trained base~\cite{akiba2025evolutionary, yang2023adamerging, yadav2023ties}. This shared origin allows for the direct application of task arithmetic~\cite{ilharco2022editing}, enabling precise manipulation of weight differences to compose capabilities. Model merging can enhance single-task performance~\cite{wortsman2022model, rame2023model, jang2024model} or be utilized in the creation of multitask models~\cite{marczak2025no, gargiulo2025task}. While merged models for multitask performance show limited promise for cross-domain compositional generalization~\cite{tam2024realistic}, we focus on explicitly reusing weights for distinct, new target tasks. 

\textbf{Singular Value Decomposition (SVD)} offers a valuable approach for parameter-efficient fine-tuning (PEFT), allowing effective modifications within the eigenspectrum of pre-trained weights~\cite{wang2024milora, balazy2024lora, peng2024sam, meng2024pissa}. While many of these strategies achieve parameter efficiency by focusing on the singular values, diverse approaches exist~\cite{lingam2024svft}. Others leverage SVD with reinforcement learning at inference time, adapting to unseen target tasks~\cite{sun2025transformer}. We introduce a unique adaptation strategy that diverges from prior work in two critical ways. First, we apply SVD to a multi-source merged model. Second, departing from the more varied heuristics seen before, our adaptation is guided exclusively by the largest singular values.

\section{Method}

\subsection{Problem statement}

Let the parameters of the base, pre-trained model be denoted by $\theta_{\text{pre}}$. We consider a set of $T$ distinct tasks. For a given task $i$, the model is fine-tuned on a corresponding dataset $D_i$. The parameters of this fine-tuned model are denoted as $\theta_i$. Finally, the parameters for a specific layer $l$ within this model are represented by $\theta_i^{(l)}$. A task vector is the element-wise difference between the parameters of a fine-tuned model and its pre-trained counterpart. Building on this concept, we define a per-layer task difference to capture these modifications with greater granularity. Denoting the parameters of the base model for layer $l$ as $\theta_{\text{pre}}^{(l)}$ and the fine-tuned parameters for task $i$ at layer $l$ as $\theta_{i}^{(l)}$, we define \textbf{task vectors} $\tau_i^{(l)}$ as $\tau_i^{(l)} = \theta_{i}^{(l)} - \theta_{\text{pre}}^{(l)}$. 
We denote matrix parameter differences (e.g., in attention layers) specifically as the \textbf{task matrix} $\Delta_i^{(l)}$ to emphasize the structure suitable for SVD, while retaining \textbf{task vector} $\tau_i^{(l)}$ as a general term interchangeably. Non-matrix parameters (e.g., biases) are averaged directly across all source tasks, similar to other works. The entire procedure, from decomposition to adaptation, is performed independently for each relevant layer in the model. For brevity, we will generally omit the layer index $(l)$. While non-parametric operations, such as activation functions, are applied during the model's forward pass, they do not have learnable weights and are therefore not represented in the task vector.


\subsection{Decomposing task matrices}

To capture the structured modifications introduced by fine-tuning, we perform a granular analysis of each task matrix, $\Delta_i$, using Singular Value Decomposition (SVD). For a given task matrix $\Delta_i$ at any generic layer, we consider its SVD:
$$
\Delta_i = \boldsymbol{U}_i \boldsymbol{\Sigma}_i \boldsymbol{V}_i^\top
$$ 
where $U_{i}\in\mathbb{R}^{m\times r_i}$ and $V_{i}\in\mathbb{R}^{n\times r_i}$ are the matrices of left and right singular vectors, respectively, and $\Sigma_{i}\in\mathbb{R}^{r_i\times r_i}$ is a diagonal matrix containing the singular values $\sigma\in\mathbb{R}^{r_i}$. The value $r_i$ denotes the rank of the matrix $\Delta_i$ and corresponds to the number of its singular components. Given a pre-trained model, parameterized by $\theta_{\text{pre}}$, and a library of $T-1$ source task matrices, $\{\Delta_i\}_{i=1}^{T-1}$, our objective is to synthesize this knowledge to effectively adapt the model for a new, unseen target task. The original training datasets for these source tasks, i.e., $\{D_1,...,D_{T-1}\}$, are not available. For the target task, we only have access to its labeled dataset, which is partitioned into a training set $D_{\text{t}}^{\text{train}}$ and a test set $D_{\text{t}}^{\text{test}}$.

\subsection{Our Two-Stage Composition Framework}

\subsubsection*{Stage 1: Extraction and Aggregation.}

Our core hypothesis (validated at Appendix Section~\ref{sec:transfer_boundaries}) is that the most useful transferable knowledge for the target task, encoded across diverse source tasks $\{\Delta_{i}\}_{i=1}^{T-1}$, is within the principal singular components, which represent the most dominant structural patterns in the parameter space. 

Therefore, for each source task matrix $\Delta_i$, we perform SVD to decompose it into a set of orthogonal components. Each component is a triplet $(\mathbf{u}_{i,j}, \tilde{\sigma}_{i,j}, \mathbf{v}_{i,j}^\top)$, where $j$ is the component index for a given task $i$, and $\tilde{\sigma}_{i,j} = \sigma_{i,j} / \|\Delta_i\|_F$ represents the singular value normalized by the Frobenius norm of its task matrix to ensure scale comparability across tasks. Consequently, we propose an aggregation strategy based on a global ranking of all such normalized components from all source task matrices. 

We then select the top-$K$ components (where $K$ depends on the total number of tasks $T$) with the highest normalized singular values to construct the transfer basis, re-indexing them as $k=1 \dots K$:
\begin{equation*}
\label{eq:transfer_basis}
\mathcal{B} = \{ (\mathbf{u}_k, \tilde{\sigma}_k, \mathbf{v}_k^\top) \}_{k=1}^K.
\end{equation*}
Finally, the merged task matrix, $\Delta_m$, is synthesized by summing the top-$K$ selected rank-one components:
\[
\Delta_m = \sum_{k=1}^{K} \mathbf{u}_k \tilde{\sigma}_k \mathbf{v}_k^\top.
\label{eq:synthesis}
\]
It is worth noting that the actual rank of $\Delta_m$ will likely be less than $K$. This occurs because the basis vectors are only orthonormal within a specific task, meaning that vectors originating from different tasks may be linear combinations of one another. By prioritizing these high-magnitude components, we aim to build a new, effective pre-trained state for any unknown downstream task. Although summing rank-one components introduces non-orthogonality, it effectively accumulates the primary directions of change from all source tasks.

\subsubsection*{Stage 2: Target Task Adaptation.}
In the second stage, the merged knowledge $\Delta_m$ is adapted to the specific target task. We define the final target task parameters $\Delta_{t}$ as a function of $\Delta_m$ and a small set of \textit{learnable parameters} $\Lambda$ that minimize the cross-entropy loss $\mathcal{L}$ on the target dataset:
$$ \Lambda^* = \underset{\Lambda}{\operatorname{argmin}} \ \mathbb{E}_{(x,y) \in D_{\text{t}}} \left[ \mathcal{L}\left(f(x;\theta_{\text{pre}} + \Delta_{\text{t}}(\Lambda)),y\right) \right] $$ For a parameter-efficient adaptation, we apply gradient-based learning exclusively to the top-$N$ singular values of $\Delta_t$, which constitute the set $\Lambda$. The singular vectors $U_t, V_t$ and the remaining singular values are kept frozen. The resulting full model parameters for the target task are $\theta_{\text{t}} = \theta_{\text{pre}} + \Delta_{\text{t}}(\Lambda)$ and the full, step-by-step process is formalized in Algorithm~\ref{alg:svd_agg} and Figure~\ref{fig:method_overview_phase_one}.

The synthesized matrix $\Delta_m$ represents a rich but intermediate consolidation of knowledge from multiple source tasks. To transform this aggregation into a computationally stable and effective basis for adaptation, we re-parameterize it using a final SVD. This procedure, $\Delta_{m} \rightarrow U_{t} \Sigma_{t} V_{t}^{\top}$, serves a dual purpose. 
First, it acts as a reconciliation step: it finds the optimal orthogonal basis ($U_t$ and $V_t$) that best approximates this accumulated field of updates in the Frobenius norm sense.
Second, it yields a new diagonal matrix $\Sigma_t$, whose values reflect the spectral energy of the aggregated update $\Delta_{m}$ and also directly serve as the isolated set of learnable parameters, $\Lambda$, for the subsequent fine-tuning.

\section{Results}

\subsection{Experimental setup}

\paragraph{Vision Benchmarks.}
To evaluate the performance, scalability, and robustness of our method in the visual domain, we benchmark it against the state-of-the-art method aTLAS and other established merging techniques integrated into our adaptation framework. The test bed is based on 21 diverse image classification tasks, including satellite imagery (EuroSAT) and fine-grained categorization (Flowers102). To ensure a fair comparison, we utilize the library of public task vectors provided by the authors of aTLAS. We employ a leave-one-out protocol: for each target task, we incrementally aggregate knowledge assets by varying the number of source task vectors from one up to the maximum of $T-1$ in a fixed, predefined sequence. By default, we use the pre-trained Vision Transformer (ViT-B-32) variant of the CLIP model~\cite{radford2021learning}, while evaluations on ViT-L-14 architectures and ablation studies with ViT-B-16 are provided in the Appendix.

\begin{figure*}[t]
    \centering
    \includegraphics[width=\linewidth]{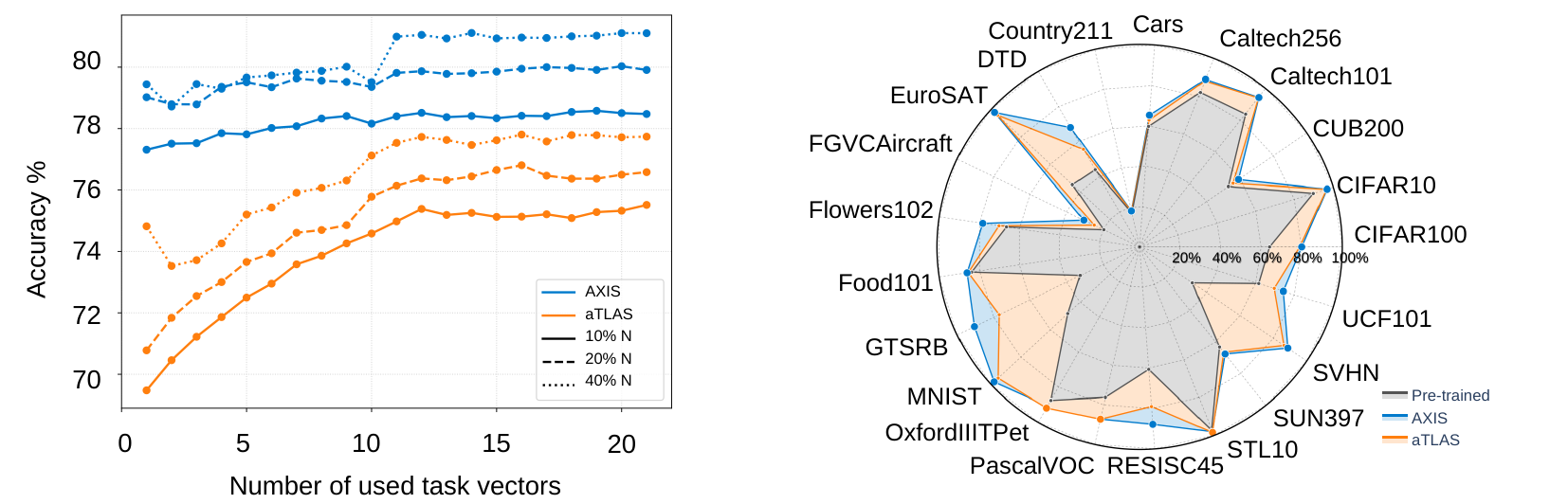}
    \caption{\textbf{Left:} Performance comparison with aTLAS varying the number of trainable parameters with the ViT-B-32 architecture. Each point represents a model configuration that was independently adapted to all target tasks. The plotted value is the mean performance across these tasks. \textbf{Right:} Detailed per-task comparison of the merged models (AXIS vs. aTLAS) utilizing 16 task vectors as a representative example. To clarify performance differences in overlapping regions, the marker of the superior method for a given task is rendered on top. Overall, AXIS achieves a higher average accuracy of 78.42\% compared to 75.13\% for aTLAS (details in Table~\ref{tab:performance_comparison_16tv}).}
    \label{fig:vit-b-32_comparison}
\end{figure*}

\begin{figure}[t]
    \centering
    \includegraphics[width=1.0\columnwidth]{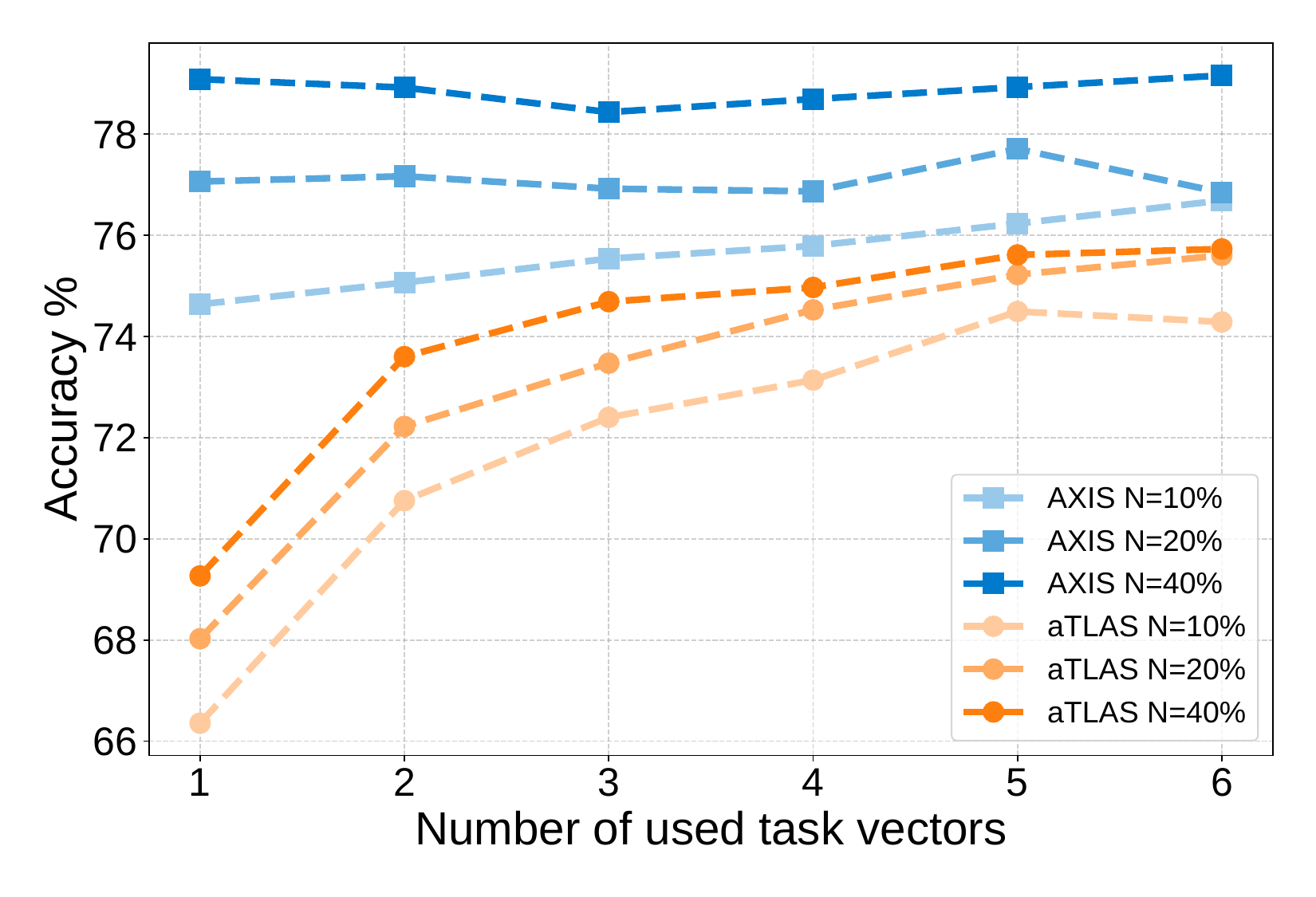}
    \caption{AXIS consistently outperforms the aTLAS baseline across seven diverse NLP benchmarks under varying source task aggregation levels and parameter budgets $N$.}
    \label{fig:t5_base_comparison}
\end{figure}

\begin{figure*}[t]
    \centering
    \includegraphics[width=\textwidth]{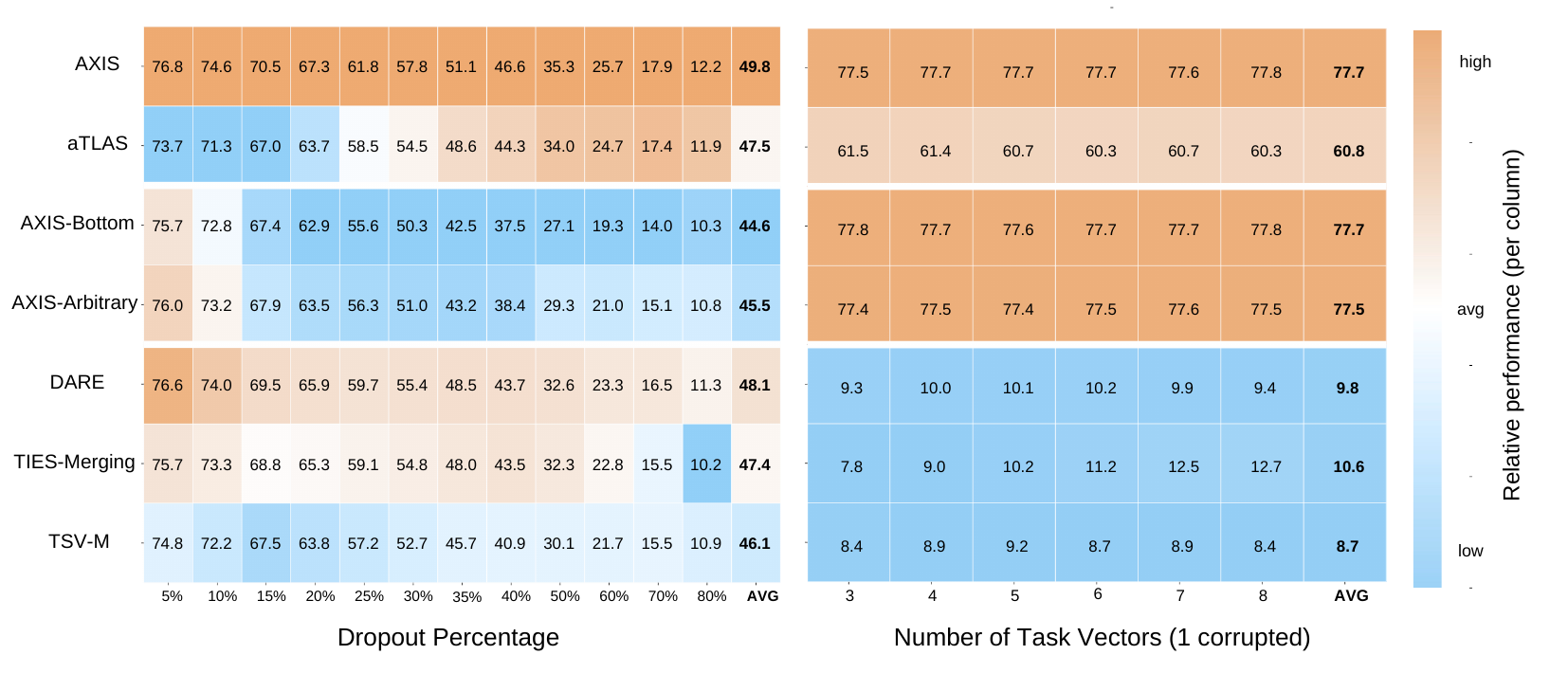}
    \caption{
        We compare AXIS primarily against the state-of-the-art aTLAS, while also including established merging methods adapted via our Stage 2 framework to serve as competitive reference points. \textbf{Left:} The heatmap illustrates the average accuracy across all target tasks. AXIS outperforms the baselines under challenging conditions where input information is partially hidden, with up to 80\% of patches masked.
        \textbf{Right:} Results averaged across all target tasks summarize the robustness to a single corrupted source task vector (varying from 3 to 8 total). Our method, AXIS, demonstrates superior resilience to this scenario compared to aTLAS and other merging methods.
    }
    \label{fig:combined_figure_robustness}
\end{figure*}

\paragraph{Language Benchmarks.}
We extend our evaluation to the Natural Language Processing (NLP) domain. We utilize the T5-Base architecture and adopt the multi-task merging protocol established in TIES-Merging~\cite{yadav2023ties}. The evaluation encompasses seven NLP datasets: 
question answering (QASC~\cite{khot2020qasc}, WikiQA~\cite{yang2015wikiqa}, and QuaRTz~\cite{tafjord2019quartz}), paraphrase identification (PAWS~\cite{zhang2019paws}), sentence
completion (Story Cloze~\cite{sharma2018tackling}), and coreference resolution (Winogrande~\cite{sakaguchi2021winogrande} and WSC~\cite{levesque2012winograd}). For NLP tasks, we average results across all valid source task vectors combinations.

All results are presented under a matched number of trainable parameters and within the range used by the aTLAS method to ensure a fair comparison. For vision tasks, we adhere to the training protocols established by the authors of aTLAS. Specifically, each adaptation runs for 10 epochs with a learning rate of $10^{-1}$ using the full standard training set. Additional experimental setup details and extensive performance comparisons are provided in the Appendix.

\subsection{Performance and Efficiency Gains over aTLAS}
\paragraph{Vision Domain} For each target task, we incrementally build the merged task vector, $\Delta_{target}$, by aggregating an increasing number of source task vectors. For example, a single model synthesized from 16 source vectors is then independently fine-tuned 21 times, once for each distinct target task as illustrated in Figure~\ref{fig:vit-b-32_comparison}. This entire process is repeated for every aggregation level as shown on the x-axis in Figure~\ref{fig:vit-b-32_comparison}, and the outcomes are averaged to produce the final performance curves. The parameter budgets $N$ of 10\%, 20\%, and 40\% are determined by the percentage of trainable singular values selected from each task matrix; their sum across all matrices results in total trainable parameter counts of approximately 3.6k, 7.3k, and 14.7k, respectively, in the ViT-B-32. The results demonstrate that our approach outperforms aTLAS across the entire spectrum of source task quantities on both the ViT-B-32 (illustrated in the right panel of Figure~\ref{fig:vit-b-32_comparison}) and larger ViT-L-14 architectures (see Figure~\ref{fig:vit-l-14_comparison} in the Appendix). 

\paragraph{Language Domain} AXIS consistently outperforms the aTLAS baseline on the T5-Base architecture across seven diverse NLP benchmarks, spanning all evaluated parameter budgets ($N \in \{10\%, 20\%, 40\%\}$) and source task aggregation levels. This demonstrates that our component-based aggregation strategy is effectively modality-agnostic and maintains its effectiveness regardless of the underlying architecture type. Performance comparisons for the language tasks are presented in Figure~\ref{fig:t5_base_comparison}.

As illustrated in the left panel of Figure~\ref{fig:param_efficiency}, AXIS consistently outperforms aTLAS across all parameter budgets. Furthermore, the noticeably smaller shaded area for AXIS indicates a lower standard deviation, highlighting that our aggregation mechanism is more stable and less sensitive to variations in the number of source task vectors used.

\paragraph{Memory and Runtime Scalability.} A key advantage of our method is its significantly lower computational overhead compared to baselines like aTLAS as shown in the right panel of Figure~\ref{fig:param_efficiency}. The memory and runtime costs of aTLAS scale near-linearly with the number of source models, as it learns a distinct coefficient for each of the $T$ source tasks across every layer and parameter partition $P$ during the fine-tuning process. This means that all source task vectors must be present in memory throughout the entire adaptation phase for a new target task. In contrast, AXIS decouples the process into two distinct stages. The first stage, knowledge aggregation, is an offline one-time operation. It efficiently processes all $T-1$ source task vectors using SVD and consolidates them into a single, fixed-size merged matrix $\Delta_m$. The subsequent fine-tuning stage operates only on this compact $\Delta_m$. As a result, the memory footprint and runtime of the adaptation phase remain constant, regardless of the number of source models initially aggregated. 

\subsection{Robustness to Noise and Sparsity in Source Parameters}

To evaluate the robustness of our method with unreliable~\cite{li2025task} or compressed~\cite{iurada2025efficient, li2025task} source task vectors, we designed two specific scenarios. The first simulates contamination from a single, low-quality source, for instance, due to training instabilities. The second scenario evaluates how effectively these approaches leverage knowledge when all source task vectors are heavily pruned. Both investigations explore the method's capacity to merge a more diverse and challenging spectrum of models, expanding its practical applicability. 

We formed aggregations of source task vectors of varying sizes, ranging from three to eight, to demonstrate the effect of a single faulty source. In each aggregation, one task vector was intentionally corrupted, while the others remained intact. The corruption was applied by adding zero-mean Gaussian noise to the weights of an original task vector. To ensure a significant level of disruption, the standard deviation of the noise was scaled to 50\% of the Frobenius norm of that task matrix ($\sigma = 0.5 \cdot ||\Delta_i||_F$). The results illustrated in the right panel of Figure~\ref{fig:combined_figure_robustness} demonstrate that while aTLAS and alternative aggregation methods refined by our adaptation stage experience some performance degradation in the presence of a corrupted source, the impact on our method is significantly less pronounced. This indicates a more robust knowledge transfer mechanism. We observe that our SVD-based selection process, by focusing on components with the highest singular values, is significantly less susceptible to the unstructured perturbations introduced into a single source vector.

To assess the robustness of our method from a compression perspective, each of the source task vectors underwent magnitude-based pruning (see Figure~\ref{fig:corruption_robustness} in Appendix). 
The subsequent analysis suggests that our approach can more effectively leverage the knowledge contained within highly sparse task vectors, showcasing a distinct advantage in utilizing compressed knowledge.  

\begin{figure}[t]
    \centering
    \includegraphics[width=\columnwidth]{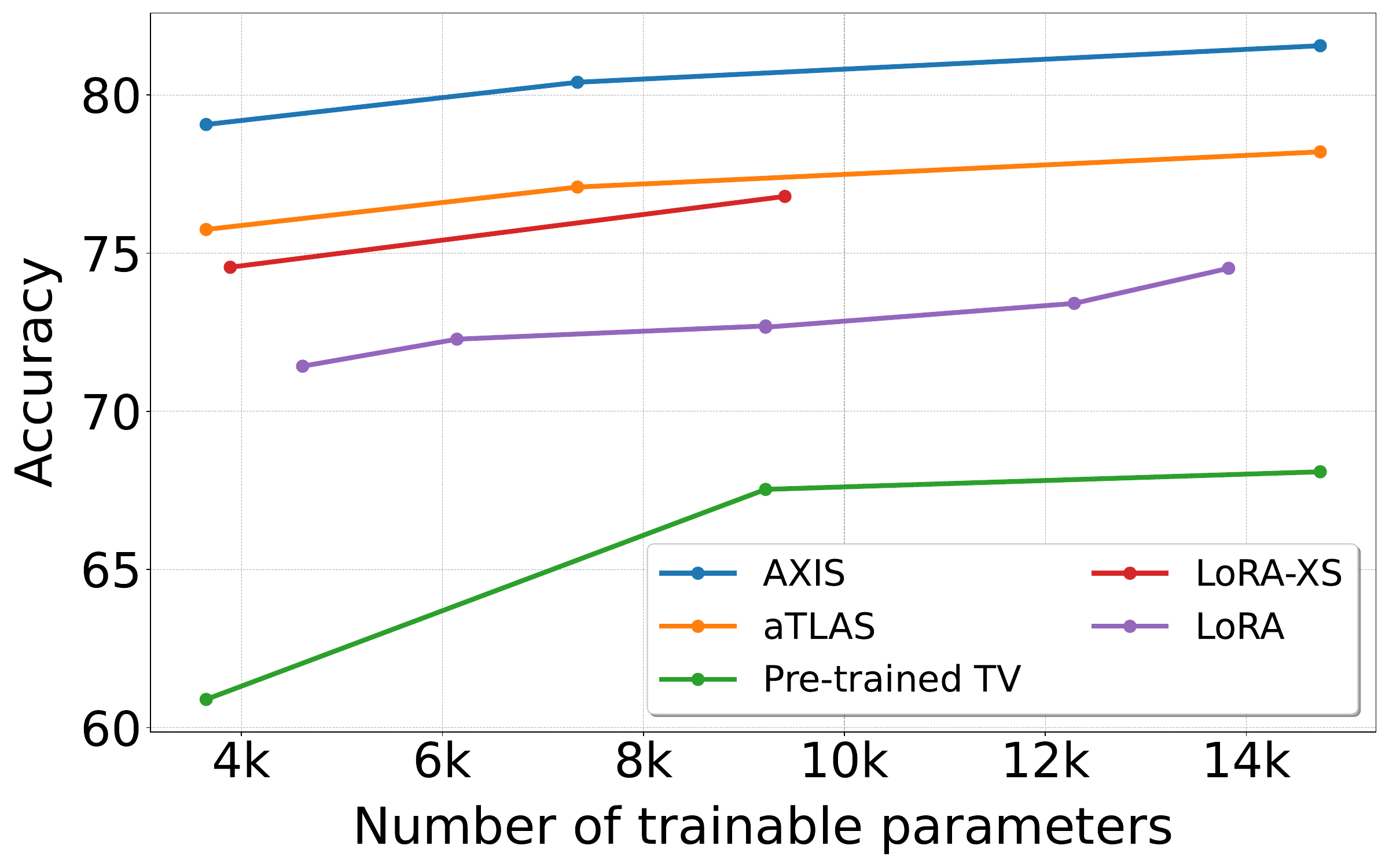}
    \caption{
    Performance comparison with competing methods, including PEFT variants. The proposed merge-and-tune paradigm in AXIS achieves a more efficient performance-parameter trade-off.
    }
    \label{fig:appendix_param_efficiency}
\end{figure}

\subsection{Robustness to Input Data Degradation}

Building on findings that merging models fine-tuned with distinct hyperparameters on the same task leads to greater stability under distribution shifts~\cite{wortsman2022model, wortsman2022robust}, we explore whether aggregating knowledge from multiple, diverse models, each fine-tuned with the same set of hyperparameters, can similarly construct a more robust representation. For this experiment, AXIS and aTLAS models were built by aggregating the complete set of $T-1$ source task vectors and fine-tuning them for each target task.

Evaluating the model on images with randomly omitted patches serves as a direct test of robustness~\cite{PaulC22} and the ability to predict with partial information~\cite{PardylKOT025}. This approach offers unique insight into the model's internal representation, as this specific form of robustness correlates less with baseline performance than other image perturbations~\cite{malik2025towards}. To ensure a fair comparison, a fixed seed guarantees that all methods are evaluated using the same masked patches for each dropout level. In the left panel of Figure~\ref{fig:combined_figure_robustness}, AXIS shows resilience when nearly all information is available, and degrades more slowly as input degradation becomes more severe. This capability is essential for real-world scenarios with incomplete data and follows prior research aimed at improving model resilience to partial visual information~\cite{liu2023patchdropout, tang2022patch} (see Table~\ref{tbl:appendix_robustness_detailed} in Appendix). Additionally, we demonstrate better robustness capabilities of AXIS than aTLAS against a set of 12 common image corruptions~\cite{hendrycks2019benchmarking} with 5 severity levels (see Figure~\ref{fig:image_corruptions} and~\ref{fig:severity_levels}).

\begin{table}[t]
    \centering
    \caption{Performance comparison with aTLAS and merging methods when followed by our Stage 2 adaptation. While the best results are obtained by AXIS, the adaptation mechanism itself is a potent and versatile tool for refining diverse multi-capability models. Crucially, while other merging baselines achieve lower, but competitive accuracy, AXIS exhibits significantly superior robustness against weights corruption (see Figure~\ref{fig:combined_figure_robustness} and~\ref{fig:corruption_robustness}). All results are averaged over 3 seeds.}
    \begin{adjustbox}{width=\columnwidth, center}
    \begin{tabular}{lccc}
    \toprule
    Method & \textbf{N = 10\%} & \textbf{N = 20\%} & \textbf{N = 40\%} \\
    \midrule
    DARE + Stage 2 & $78.09 \pm 0.06$ & $79.69 \pm 0.04$ & $80.77 \pm 0.09$ \\
    TIES + Stage 2 & $77.39 \pm 0.03$ & $78.99 \pm 0.05$ & $80.27 \pm 0.05$ \\
    TSV-M + Stage 2 & $76.41 \pm 0.05$ & $78.69 \pm 0.07$ & $80.41 \pm 0.11$ \\
    \midrule
    aTLAS & $75.50 \pm 0.03$ & $75.93 \pm 0.44$ & $77.66 \pm 0.05$ \\
    \textbf{AXIS} & \bm{$78.46 \pm 0.04$} & \bm{$79.93 \pm 0.11$} & \bm{$81.13 \pm 0.07$} \\
    \bottomrule
    \end{tabular}
    \end{adjustbox}
    \label{tab:merging-methods-comparison}
\end{table}

\subsection{Comparison with PEFT and Merging}
\label{sec:peft_merg_comparision}

To demonstrate the advantages of our approach, we compare it with different fine-tuning methods, in particular with PEFT methods. This includes the widely-adopted LoRA~\cite{hu2022lora} and its enhanced variant LoRA-XS~\cite{balazy2024lora}. Additionally, we try to further adapt the pre-trained weights as a single task vector. As Fig~\ref{fig:appendix_param_efficiency} shows, our method efficiently outperforms these techniques, effectively reusing already fine-tuned weights.

We further ask the question whether a general, multi-task model serves as an effective knowledge base for our Stage 2 adaptation. To test this hypothesis, we substitute our AXIS aggregation with several established multi-task merging techniques, such as DARE~\cite{yu2024language}, TIES-Merging~\cite{yadav2023ties} and TSV-M~\cite{gargiulo2025task}, treating their merged weights as alternative initializations. As the results in Table~\ref{tab:merging-methods-comparison} demonstrate, these multi-task models indeed form a potent foundation for our adaptation mechanism, albeit slightly below the performance of the AXIS method. This suggests that Stage 2 is not rigidly dependent on a single aggregation method but can effectively refine knowledge from various merged, multi-capability models. Crucially, these alternative merging baselines lack the structural robustness inherent in our approach, as illustrated in Figure~\ref{fig:combined_figure_robustness}. They exhibit significant performance degradation in realistic scenarios involving partial input information or corrupted source models. 

\section{Ablation Study}

\subsection{Components Selection}
\label{sec:component_selection_main}
We base our selection strategy on the empirical finding that the most transferable knowledge is encapsulated within the principal singular components of the source tasks. As our analysis reveals, components with the highest singular values consistently align best with the ground-truth target task updates, while transferability sharply declines for lower-ranked components. Validation of this phenomenon, including energy distribution analysis and per-component performance plots, is provided in Appendix Section~\ref{sec:svd_components_analysis} (see Figure~\ref{fig:cal_top_k_exp} and~\ref{fig:oracle_top_k_exp}). 

\subsection{Robustness to Component Selection Size}

\begin{figure}[tb]
    \centering
    \includegraphics[width=1.0\columnwidth]{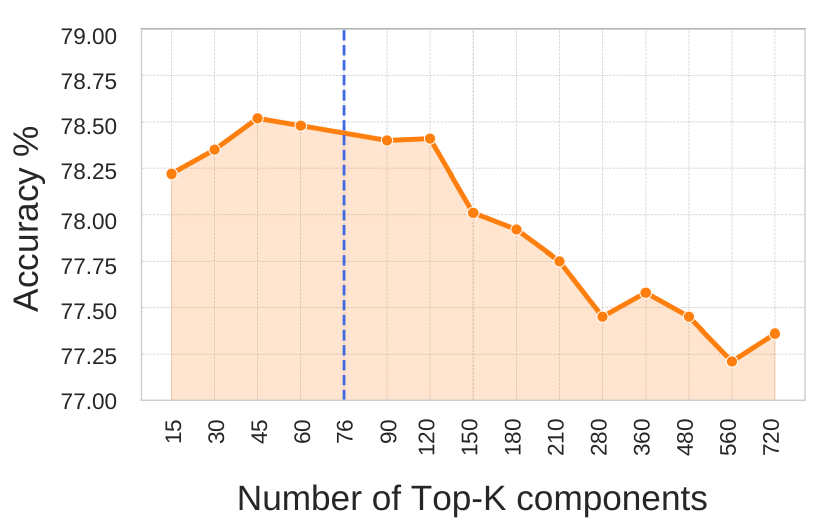}
    \caption{Performance sensitivity to the number of aggregated components $K$. The blue vertical line marks our default setting of $K=76$. AXIS is robust to parameter changes as the absolute performance variance remains minimal ($<1.5\%$) across a wide range of $K$ values.}
    \label{fig:topK_components}
\end{figure}

To assess the sensitivity of our method to the size of the transfer basis, we conducted an ablation study on the number of selected components $K$. This sole hyperparameter directly controls the dimensionality of the aggregated knowledge consolidated into the merged task matrix, $\Delta_m$. In this experiment, we varied the value of $K$ used in our \textit{top components} aggregation strategy, where components from all source tasks are globally ranked by their singular values. 
Our default choice of $K=76$ (approximating 10\% of each layer's rank) proves to be a robust heuristic. The plot demonstrates that performance remains high, with the drop being less than 1.5\% even for large $K$ (Figure~\ref{fig:topK_components}). Overall, we find that limiting $K$ to be less than 20\% of the layer's full rank yields robust results. We hypothesize that including additional components (the spectral tail) may introduce more task-specific details, which are not necessarily important for the target task (see Figure~\ref{fig:cal_top_k_exp} and~\ref{fig:oracle_top_k_exp}).

\subsection{Components Selection Strategy} 

\begin{figure}[t]
    \centering
    \captionof{table}{Performance comparison of different SVD component selection strategies within the AXIS framework across varying parameter budgets ($N$), demonstrating their comparable effectiveness. However, the choice of components is crucial for ensuring the resilience of the model, as illustrated in Figure~\ref{fig:combined_figure_robustness} and~\ref{fig:top_axis_common_corruption}. Results are averaged over three random seeds.}
    \begin{adjustbox}{width=\linewidth,center}
    \begin{tabular}{rcccc}
    \toprule
    \textbf{N (\%)} & \textbf{Top} & \textbf{Arbitrary} & \textbf{Bottom} \\
    \midrule
    10  & $78.46 \pm 0.04$ & $77.83 \pm 0.04$ & $77.56 \pm 0.02$\\
    20  & $79.93 \pm 0.11$ & $79.79 \pm 0.03$ & $79.81 \pm 0.05$\\
    40  & $81.13 \pm 0.07$ & $81.17 \pm 0.08$ & $81.13 \pm 0.04$\\
    \bottomrule
    \end{tabular}
    \end{adjustbox}
    \label{tab:axis_variants_comparison_final}
\end{figure}

To evaluate the quality of component aggregation, we test three selection criteria from a global pool of all aggregated SVD components. We compare the impact of selecting components with the highest singular values (\textit{top components}), the lowest (\textit{bottom components}), and those chosen arbitrarily (\textit{arbitrary components}). The results of this comparison are presented in Table~\ref{tab:axis_variants_comparison_final}, which indicates that the \textit{top components} strategy yields the best performance. While selecting the \textit{top components} yields the highest accuracy, this advantage is most pronounced at lower parameter budgets. As the number of trainable parameters increases, the performance of all three strategies converges, suggesting that the importance of the initial component selection decreases as the model is given more trainable parameters. However, robustness remains a key differentiator (see Figure~\ref{fig:combined_figure_robustness}). Alternative selection strategies exhibit significantly lower resilience against both partial input information and common corruptions (see also Figure~\ref{fig:top_axis_common_corruption} in the Appendix).





\subsection{Additional Analyses}

Due to space constraints, the Appendix includes complementary analyses not covered in the main text. First, we demonstrate that AXIS scales effectively to $N$=100\% of its parameter budget, achieving parity with full fine-tuning (see Section~\ref{sec:axis_scalability}). Second, we empirically justify the second SVD re-parameterization, showing it is critical for stabilizing the transfer basis and decoupling learnable parameters (see Section~\ref{sec:stabilizing_tranfer_basis}). Third, we validate our Stage 2 adaptation strategy, confirming that optimizing singular values outperforms alternative strategies such as fine-tuning singular vectors (see Section~\ref{sec:validation_of_adaptation_flex}). Crucially, we investigate transfer boundaries, presenting promising evidence that AXIS operates effectively even beyond shared initializations, enabling knowledge transfer across minor architectural variations and different model scales (see Section~\ref{sec:transfer_boundaries}).

\section{Conclusion}

We presented AXIS, an efficient and scalable framework that addresses multi-source knowledge transfer in vision and language domains through the spectral extraction, aggregation, and adaptation of parameters for the target task. Our approach overcomes the scalability bottlenecks associated with aggregating numerous source models and high parameter counts. By maintaining a constant runtime and memory footprint, AXIS resolves these computational constraints. The framework enables efficient final adaptation while demonstrating notable intrinsic robustness to degradations at both the parameter and input levels. 



\section*{Impact Statement}
This paper presents work whose goal is to advance the field of Machine Learning. There are many potential societal consequences of our work, none which we feel must be specifically highlighted here.

\bibliography{my_paper}
\bibliographystyle{icml2026}

\newpage
\appendix
\onecolumn

\section*{Overview}

\begin{itemize}
    \item \textbf{Evaluation protocol} is described in Section~\ref{sec:eval_protocol}. 
    \item \textbf{Performance on ViT-L-14} is presented in Section~\ref{sec:vit_l_14}.
    \item \textbf{Performance on T5-Base} is detailed in Section~\ref{sec:t5_base}.
    \item \textbf{AXIS scalability with N} and comparison with full-parameter fine-tuning are analyzed in Section~\ref{sec:axis_scalability}.
    \item \textbf{Stabilizing the transfer basis} via final SVD is discussed in Section~\ref{sec:stabilizing_tranfer_basis}.
    \item \textbf{Validation of adaptation flexibility} is analyzed in Section~\ref{sec:validation_of_adaptation_flex}.
    \item \textbf{Investigating transfer boundaries} is explored in Section~\ref{sec:transfer_boundaries}.
    \item \textbf{Transferability by SVD components} is justified in Section~\ref{sec:svd_components_analysis}.
    \item \textbf{Incremental knowledge aggregation} is introduced in Section~\ref{sec:incremental}.
    \item \textbf{In-depth robustness analyses} including corruptions and data availability are in Section~\ref{sec:robustness}.
    \item \textbf{Component selection} ablation studies are presented in Section~\ref{sec:component_selection}.
    \item \textbf{Detailed main results} tables are provided in Section~\ref{sec:detailed_results}.
\end{itemize}

\section{Evaluation protocol}
\label{sec:eval_protocol}

\subsection{Vision Models}
\begin{table}[h]
\centering
\caption{Overview of the datasets, data splits, and training epochs used in the experiments, reporting the fine-tuning accuracy for ViT-B-32 and ViT-L-14 models.}
\resizebox{0.85\columnwidth}{!}{%
\begin{tabular}{@{}lcrrrccc@{}}
\toprule
\multirow{2}{*}{\textbf{Dataset}} & \multirow{2}{*}{\textbf{Classes}} & \multicolumn{3}{c}{\textbf{Splits}} & \multirow{2}{*}{\textbf{Epochs}} & \multicolumn{2}{c}{\textbf{Fine-tuned accuracy (\%)}} \\ \cmidrule(lr){3-5} \cmidrule(l){7-8}
 & & \multicolumn{1}{c}{train} & \multicolumn{1}{c}{val} & \multicolumn{1}{c}{test} & & ViT-B/32 & ViT-L/14 \\ \midrule
Cars & 196 & 7,330 & 814 & 8,041 & 35 & 78.26 & 91.67 \\
DTD & 47 & 3,384 & 376 & 1,880 & 76 & 78.94 & 84.73 \\
EuroSAT & 10 & 21,600 & 2,700 & 2,700 & 12 & 98.89 & 99.81 \\
GTSRB & 43 & 23,976 & 2,664 & 12,630 & 11 & 99.14 & 99.30 \\
MNIST & 10 & 55,000 & 5,000 & 10,000 & 5 & 99.65 & 99.77 \\
RESISC45 & 45 & 17,010 & 1,890 & 6,300 & 15 & 95.94 & 97.14 \\
SUN397 & 397 & 17,865 & 1,985 & 19,850 & 14 & 75.40 & 81.98 \\
SVHN & 10 & 68,257 & 5,000 & 26,032 & 4 & 97.38 & 97.97 \\
CIFAR10 & 10 & 45,000 & 5,000 & 10,000 & 5 & 98.05 & 99.22 \\
CIFAR100 & 100 & 45,000 & 5,000 & 10,000 & 6 & 89.09 & 93.01 \\
ImageNet & 1,000 & 1,276,167 & 5,000 & 50,000 & 10 & 76.41 & 85.52 \\
STL10 & 10 & 4,500 & 500 & 8,000 & 4 & 98.55 & 99.62 \\
Food101 & 101 & 70,750 & 5,000 & 25,250 & 15 & 88.68 & 95.37 \\
Caltech101 & 101 & 6,941 & 694 & 1,736 & 10 & 94.41 & 94.82 \\
Caltech256 & 257 & 22,037 & 2,448 & 6,122 & 8 & 92.60 & 97.17 \\
FGVCAircraft & 100 & 3,334 & 3,333 & 3,333 & 60 & 40.65 & 68.11 \\
Flowers102 & 102 & 1,020 & 1,020 & 6,149 & 40 & 90.08 & 97.84 \\
OxfordIIITPet & 37 & 3,312 & 368 & 3,669 & 5 & 92.15 & 95.91 \\
CUB200 & 200 & 5,395 & 599 & 5,794 & 20 & 73.56 & 86.35 \\
PascalVOC & 20 & 7,844 & 7,818 & 14,976 & 10 & 88.42 & 92.05 \\
Country211 & 211 & 31,650 & 10,550 & 21,100 & 15 & 21.99 & 38.06 \\
UCF101 & 101 & 7,639 & 1,898 & 3,783 & 20 & 85.01 & 92.55 \\ \bottomrule
\end{tabular}
}
\label{tab:full_finetuning_accuracy}
\end{table}

We adopt the comprehensive benchmark, publicly released task vectors, and training protocols established by the authors of aTLAS. Their framework provides task vectors obtained by fine-tuning the pre-trained CLIP~\cite{radford2021learning} model on distinct image recognition datasets: Stanford Cars~\cite{krause20133d}, DTD~\cite{cimpoi2014describing}, EuroSAT~\cite{helber2019eurosat}, GTSRB~\cite{stallkamp2011german}, MNIST~\cite{lecun1998mnist}, RESISC45~\cite{cheng2017remote}, SUN397~\cite{xiao2016sun}, SVHN~\cite{netzer2011reading}, CIFAR10~\cite{krizhevsky2009learning}, CIFAR100~\cite{krizhevsky2009learning}, ImageNet~\cite{russakovsky2015imagenet}, STL10~\cite{coates2011analysis}, Food101~\cite{bossard2014food}, Caltech101~\cite{fei2006one}, Caltech256~\cite{griffin2007caltech}, FGVCAircraft~\cite{maji2013fine}, Flowers102~\cite{nilsback2008automated}, Oxford Pets~\cite{parkhi2012cats}, CUB200~\cite{welinder2010caltech}, PascalVOC~\cite{everingham2015pascal}, Country211~\cite{radford2021learning}, and UCF101~\cite{soomro2012ucf101}. The original fine-tuning for these vectors was performed using the AdamW optimizer~\cite{loshchilov2017decoupled} with a learning rate of $10^{-5}$, a batch size of 128, and a weight decay of 0.1 for the ViT-B-32 architecture. Table~\ref{tab:full_finetuning_accuracy} provides dataset details, their corresponding hyperparameters, and the fine-tuning accuracy achieved with full fine-tuning.

During the target task adaptation stage, we fine-tune the merged model for each dataset independently, using the same hyperparameters as the aTLAS baseline (each adaptation runs for 10 epochs with a learning rate of $10^{-1}$). The batch size is adjusted based on the model architecture: 64 for the ViT-B-32 and ViT-B-16 model and 128 for the larger ViT-L-14 model. For the ViT-L-14 architecture, both methods originally use two steps of gradient accumulation. To ensure a controlled and reproducible evaluation provided by aTLAS, the source task vectors are aggregated incrementally in a fixed, pre-defined sequence. The order of aggregation is as follows: Cars, DTD, EuroSAT, GTSRB, MNIST, RESISC45, SUN397, SVHN, CIFAR10, CIFAR100, ImageNet, STL10, Food101, Caltech101, Caltech256, FGVCAircraft, Flowers102, OxfordIIITPet, CUB200, PascalVOC, Country211, and UCF101.

\subsection{Computational Environment}
All experiments were conducted within a high-performance computing (HPC) cluster equipped with a heterogeneous GPU environment. The available resources included partitions with NVIDIA RTX 4090, NVIDIA V100, and NVIDIA A100 GPUs. The results reported in this paper, generated using the ViT-L-14 architecture, were obtained with nodes equipped with NVIDIA A100-SXM4-80GB GPUs. Our software stack was built upon the CUDA 12.2 toolkit with NVIDIA driver version 535.183.01.

\section{Performance on ViT-L-14}
\label{sec:vit_l_14}
\begin{figure}[h]
    \centering
    \includegraphics[width=0.8\columnwidth]{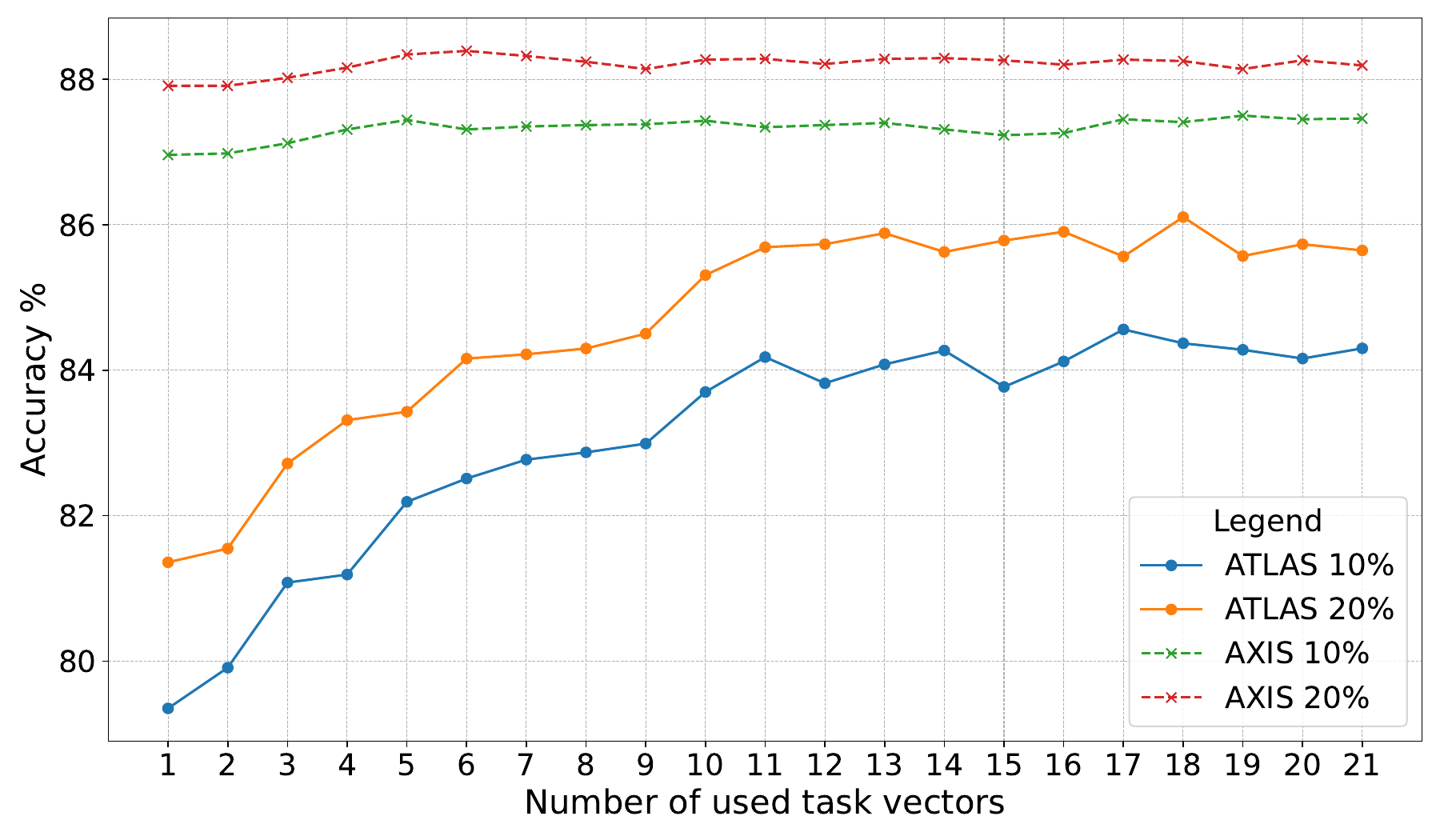}
    \caption{AXIS outperforms aTLAS on the ViT-L-14 architecture with $N$ = 10\% and $N$ = 20\% of trainable singular values. Each point is the mean accuracy across 21 independently evaluated target tasks. The plot illustrates the accuracy gain as the number of aggregated source tasks increases.}
    \label{fig:vit-l-14_comparison}
\end{figure}

To validate the scalability and effectiveness of our approach on larger models, we replicated our experiments using the ViT-L-14 architecture. The results demonstrate the advantages of the AXIS framework. The performance comparison for the $N=10\%$ and $N=20\%$ parameter budget is illustrated in Figure~\ref{fig:vit-l-14_comparison}, where AXIS consistently outperforms aTLAS as the number of aggregated source tasks increases.

\section{Performance on T5-Base}
\label{sec:t5_base}

\begin{figure}[H]
    \centering
    \includegraphics[width=0.7\columnwidth]{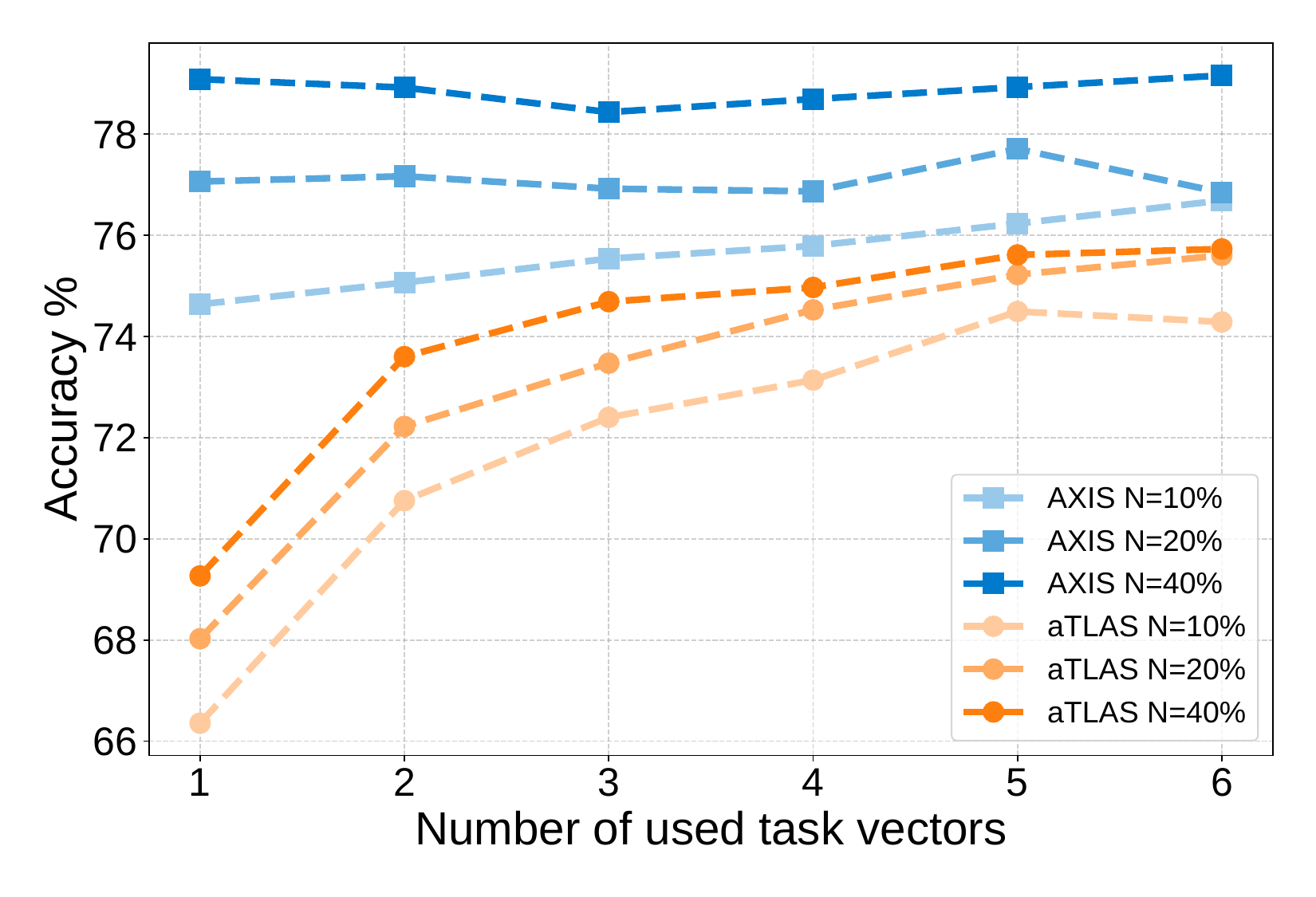}
    \caption{AXIS consistently outperforms the aTLAS baseline across seven diverse NLP benchmarks under varying source task aggregation levels and parameter budgets.}
    \label{fig:t5_base_comparison_appendix}
\end{figure}

We extend the evaluation of the AXIS framework to the language domain using the T5-base architecture, adopting the multi-task merging protocol established in TIES-Merging. The evaluation encompasses seven NLP datasets: 
question answering (QASC~\cite{khot2020qasc}, WikiQA~\cite{yang2015wikiqa}, and QuaRTz~\cite{tafjord2019quartz}), paraphrase identification (PAWS~\cite{zhang2019paws}), sentence
completion (Story Cloze~\cite{sharma2018tackling}), and coreference resolution (Winogrande~\cite{sakaguchi2021winogrande} and WSC~\cite{levesque2012winograd}).

Figure~\ref{fig:t5_base_comparison_appendix} reports the average performance for an increasing number of aggregated source models, denoted by $s$ (ranging from $1$ to $T-1$). To ensure statistical robustness, we performed an exhaustive evaluation of all valid source subsets. Specifically, for any given target task with a pool of $T-1=6$ available sources, we averaged the results across all possible combinations for each subset size $s$. This entailed computing the mean performance over $\binom{6}{1}=6$, $\binom{6}{2}=15$, $\binom{6}{3}=20$, $\binom{6}{4}=15$, $\binom{6}{5}=6$, and $\binom{6}{6}=1$ distinct source combinations per target task. The final results are averaged over all target tasks. We benchmark AXIS across three distinct trainable parameter budgets, defined by the percentage of fine-tuned singular values ($N \in \{10\%, 20\%, 40\%\}$). For a direct and fair comparison of efficiency, the aTLAS baseline was evaluated using a matching budget of trainable parameters. The empirical results demonstrate that AXIS consistently outperforms the baseline across all aggregation levels, confirming that the method's efficacy in multi-source knowledge transfer generalizes beyond the vision domain.

\section{AXIS scalability with $N$}
\label{sec:axis_scalability}

\begin{table}[H]
\centering
\caption{Comparison of AXIS with $N$=100\% with the full-parameter fine-tuning in ViT-B-32 architecture.}
\begin{tabular}{lcc}
Dataset Name & Fully Fine-tuned & AXIS \\
\midrule
CIFAR10 & 98.05 & 97.67 \\
CIFAR100 & 89.09 & 84.78 \\
CUB200 & 73.56 & 66.30 \\
Caltech101 & 76.41 & 94.89 \\
Caltech256 & 92.60 & 90.64 \\
Cars & 78.26 & 72.33 \\
Country211 & 21.99 & 19.79 \\
DTD & 78.94 & 74.49 \\
EuroSAT & 98.89 & 98.65 \\
FGVCAircraft & 40.65 & 44.74 \\
Flowers102 & 90.08 & 85.92 \\
Food101 & 94.41 & 87.88 \\
GTSRB & 99.14 & 95.92 \\
MNIST & 99.65 & 99.10 \\
OxfordIIITPet & 92.15 & 90.22 \\
PascalVOC & 88.42 & 86.79 \\
RESISC45 & 95.94 & 93.53 \\
STL10 & 88.68 & 97.09 \\
SUN397 & 75.40 & 71.86 \\
SVHN & 97.38 & 94.67 \\
UCF101 & 85.01 & 81.01 \\
\midrule
Average & 83.56 & 82.30 \\
\midrule
Train epochs  & varies from 4 to 76 & 10 \\
\end{tabular}
\label{tab:comparison_full_parameters_axis}
\end{table}

\begin{figure}[H]
    \centering
    \includegraphics[width=0.5\columnwidth]{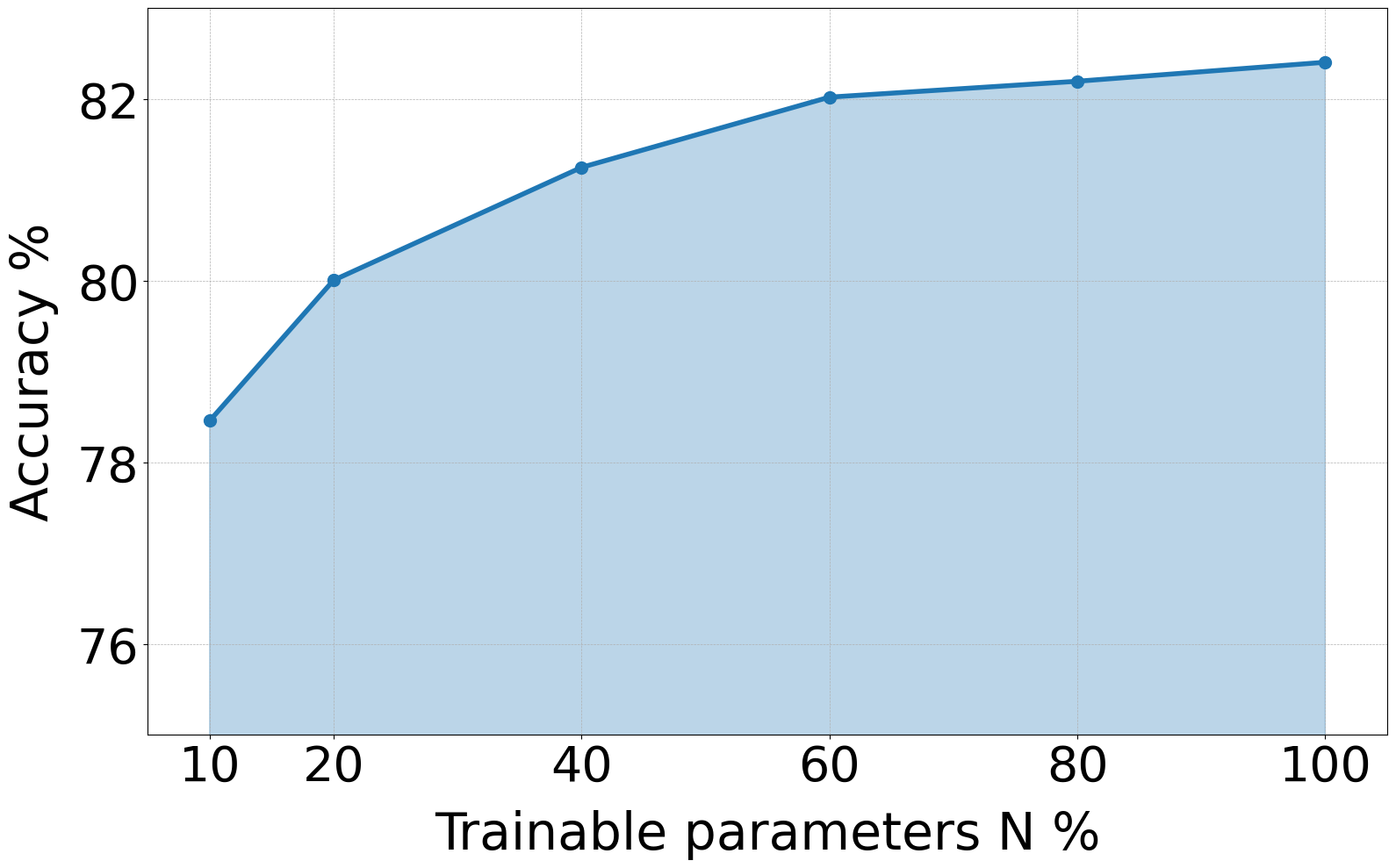}
    \caption{AXIS scales consistently with the number of trained parameters ($N$\%), showing improved performance as $N$ increases, with gains tapering off beyond 60\%.}
    \label{fig:single_point_experiment}
\end{figure}

The full-parameter performance of AXIS was evaluated by setting the singular value budget to $N$=100\%, which corresponds to fine-tuning all the singular values in each AXIS task matrix. As illustrated in Table~\ref{tab:comparison_full_parameters_axis}, the method achieved an average accuracy of 82.30\%, which closely approaches the standard full fine-tuning baseline 83.56\%. Crucially, this near-equivalent performance was attained using a fixed, significantly shorter training schedule of 10 epochs across all datasets, in contrast to the baseline's optimized, dataset-specific training that required up to 76 epochs. AXIS approaches the performance of full fine-tuning while maintaining superior computational efficiency. Additionally, we evaluate how scaling the number of trainable parameters ($N$) affects model performance (Figure~\ref{fig:single_point_experiment}). The improvements begin to diminish once $N$ exceeds 60\%. 

\section{Stabilizing the transfer basis}
\label{sec:stabilizing_tranfer_basis}

\begin{figure}[h]
    \centering
    \includegraphics[width=0.7\linewidth]{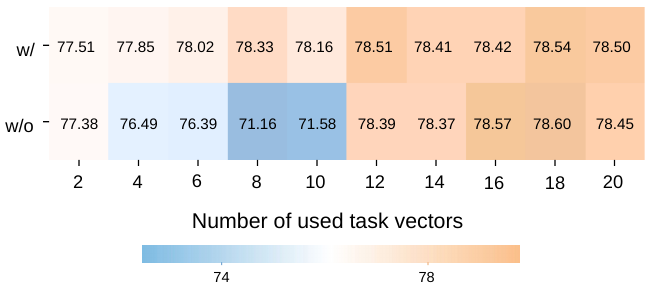}
    \caption{Skipping the final SVD orthogonalization results in a decline in performance, especially when combining a moderate number of task vectors.}
    \label{fig:final_svd}
\end{figure}

Instead of performing the final SVD re-parameterization, the layer's weights were reconstructed directly from the aggregated components $\Delta_m$. For our primary strategy of top component selection, this omission results in significant performance degradation when a moderate number of task vectors are aggregated (Figure~\ref{fig:final_svd}).

To empirically validate the importance of the final SVD re-parameterization, as discussed in the main text, we conduct a detailed ablation study. Table~\ref{tab:svd_ablation_transposed} presents a performance comparison of four different component aggregation strategies, each evaluated with and without the final SVD step.

The omission of the final SVD step is particularly detrimental to the top components strategy, resulting in a significant performance drop (e.g., over eight percentage points when aggregating 9 task vectors). In contrast, strategies based on bottom or average components exhibit significantly higher resilience to this omission. We hypothesize that two related factors drive this phenomenon. First, the top components, representing high-magnitude task-specific knowledge, likely exhibit more substantial destructive interference when their non-orthogonal vectors are directly summed. Second, this instability may be amplified during the fine-tuning process. Without a shared orthogonal basis provided by the final SVD, the learnable parameters (a subset of singular values) may conflict with the frozen components, as their underlying vectors are not decorrelated. This could lead to an unstable optimization process where adjustments to learnable components negatively interfere with the knowledge stored in the frozen ones. The relative stability of the bottom components strategy suggests that the interference from low-magnitude components is negligible, making the final orthogonalization beneficial but not as critical.

\begin{table}[H]
    \centering
    \caption{Performance comparison of different aggregation strategies with and without the final SVD step, across a varying number of aggregated task vectors and different component selection strategies.}
    \begin{adjustbox}{width=0.8\textwidth,center}
    \begin{tabular}{c c c c c c c c c}
        \toprule
        \multirow{2}{*}{\shortstack{\textbf{Aggregated}\\\textbf{Task Vectors}}} & \multicolumn{2}{c}{\textbf{Top}} & \multicolumn{2}{c}{\textbf{Bottom}} & \multicolumn{2}{c}{\textbf{Average top}} & \multicolumn{2}{c}{\textbf{Average bottom}} \\
        \cmidrule(lr){2-3} \cmidrule(lr){4-5} \cmidrule(lr){6-7} \cmidrule(lr){8-9}
        & \textbf{SVD \checkmark} & \textbf{SVD \tikzxmark} & \textbf{SVD \checkmark} & \textbf{SVD \tikzxmark} & \textbf{SVD \checkmark} & \textbf{SVD \tikzxmark} & \textbf{SVD \checkmark} & \textbf{SVD \tikzxmark} \\
        \midrule
        1 & 77.31 & 77.35 & 77.63 & 77.57 & 77.42 & 77.42 & 77.62 & 77.25 \\
        2 & 77.51 & 77.38 & 77.80 & 77.65 & 77.42 & 77.41 & 77.58 & 77.23 \\
        3 & 77.52 & 76.36 & 77.83 & 77.80 & 77.41 & 77.37 & 77.74 & 77.29 \\
        4 & 77.85 & 76.49 & 77.79 & 77.61 & 77.74 & 77.75 & 77.74 & 77.33 \\
        5 & 77.81 & 76.56 & 77.54 & 77.75 & 77.86 & 77.83 & 77.82 & 77.14 \\
        6 & 78.02 & 76.39 & 77.88 & 77.85 & 77.85 & 77.95 & 77.93 & 77.35 \\
        7 & 78.08 & 76.40 & 77.72 & 77.85 & 78.14 & 78.20 & 78.13 & 77.51 \\
        8 & 78.33 & 71.16 & 77.91 & 77.84 & 77.96 & 77.98 & 77.99 & 77.53 \\
        9 & 78.41 & 69.85 & 77.92 & 77.85 & 78.09 & 78.13 & 77.88 & 77.64 \\
        10 & 78.16 & 71.58 & 77.87 & 77.84 & 78.36 & 78.24 & 77.96 & 77.50 \\
        11 & 78.40 & 78.52 & 77.92 & 77.84 & 78.39 & 78.42 & 78.07 & 77.49 \\
        12 & 78.51 & 78.39 & 77.72 & 77.80 & 78.26 & 78.28 & 78.10 & 77.34 \\
        13 & 78.37 & 78.49 & 77.80 & 77.77 & 78.30 & 78.38 & 77.87 & 77.68 \\
        14 & 78.41 & 78.37 & 77.71 & 77.72 & 78.36 & 78.20 & 77.86 & 77.54 \\
        15 & 78.34 & 78.53 & 77.64 & 77.66 & 78.25 & 78.21 & 78.01 & 77.52 \\
        16 & 78.42 & 78.57 & 77.66 & 77.76 & 78.28 & 78.29 & 78.02 & 77.36 \\
        17 & 78.41 & 78.51 & 77.84 & 77.77 & 78.33 & 78.28 & 78.13 & 77.56 \\
        18 & 78.54 & 78.60 & 77.88 & 77.76 & 78.44 & 78.30 & 78.20 & 77.70 \\
        19 & 78.58 & 78.50 & 77.80 & 77.65 & 78.20 & 78.31 & 78.16 & 77.55 \\
        20 & 78.50 & 78.45 & 77.89 & 77.79 & 78.19 & 78.18 & 78.05 & 77.64 \\
        21 & 78.48 & 78.49 & 77.75 & 77.78 & 78.23 & 78.27 & 78.08 & 77.37 \\
        \bottomrule
    \end{tabular}
    \end{adjustbox}
    \label{tab:svd_ablation_transposed}
\end{table}

\section{Validation of adaptation flexibility}
\label{sec:validation_of_adaptation_flex}

To validate the hypothesis that tuning principal singular values does not limit adaptation flexibility, we conducted two comparative experiments while maintaining an identical budget of trainable parameters ($N$):
\begin{enumerate}
    \item \textbf{Values vs. Vectors:} We compared our method (tuning values, freezing vectors) against an alternative approach where the singular values are frozen, and a random subset of parameters within the singular vectors is fine-tuned.
    \item \textbf{Diagonal vs. Random Elements:} We investigated the structural importance of the singular value matrix, $\Sigma_t$. We train randomly selected elements of $\Sigma_t$ (allowing for off-diagonal interactions).
\end{enumerate}

In both cases, the proposed approach outperformed the alternatives as seen in Table~\ref{tab:abs_singular_values_flexibility}. These results confirm that recalibrating the importance of stable basis vectors via the principal singular values is superior to directly modifying the vectors or introducing arbitrary off-diagonal terms.

\begin{table}[H]
\centering
\caption{Comparison of adaptation strategies of Stage 2 across all target tasks for the ViT-B-16 model under different fine-tuning strategies. Our method outperforms alternative tuning strategies while maintaining an identical parameter budget.}
\resizebox{0.7\textwidth}{!}{
\begin{tabular}{lccc}
\toprule
\textbf{Dataset} & \textbf{AXIS (FT $\Sigma_t$ Diag)} & \textbf{FT Singular Vectors} & \textbf{FT $\Sigma_t$ Random} \\
\midrule
CIFAR10 & 98.29\% & 92.05\% & 97.92\% \\
CIFAR100 & 86.61\% & 66.96\% & 84.31\% \\
CUB200 & 70.85\% & 34.93\% & 65.17\% \\
Caltech101 & 96.89\% & 81.11\% & 96.14\% \\
Caltech256 & 93.24\% & 78.05\% & 91.56\% \\
Cars & 79.22\% & 48.35\% & 75.04\% \\
Country211 & 24.84\% & 13.36\% & 21.01\% \\
DTD & 73.35\% & 31.28\% & 68.40\% \\
EuroSAT & 98.48\% & 72.00\% & 98.78\% \\
FGVCAircraft & 40.11\% & 14.10\% & 38.07\% \\
Flowers102 & 86.03\% & 29.65\% & 72.08\% \\
Food101 & 91.94\% & 81.29\% & 90.19\% \\
GTSRB & 95.30\% & 43.95\% & 94.75\% \\
MNIST & 99.04\% & 85.88\% & 99.05\% \\
OxfordIIITPet & 94.85\% & 77.00\% & 93.59\% \\
PascalVOC & 89.82\% & 45.83\% & 88.43\% \\
RESISC45 & 94.08\% & 54.43\% & 93.10\% \\
STL10 & 99.31\% & 89.78\% & 98.78\% \\
SUN397 & 74.01\% & 46.19\% & 70.92\% \\
SVHN & 95.19\% & 77.77\% & 94.81\% \\
UCF101 & 84.93\% & 41.69\% & 80.49\% \\
\midrule
\textbf{Average} & \textbf{84.11\%} & \textbf{57.41\%} & \textbf{81.55\%} \\
\bottomrule
\end{tabular}%
}
\label{tab:abs_singular_values_flexibility}
\end{table}

\section{Investigating transfer boundaries}
\label{sec:transfer_boundaries}

While our primary objective is the direct comparison against the aTLAS baseline, necessitating a shared initialization, we design entirely new experiments to test the practical boundaries of the AXIS framework. 

Notably, our results point to a phenomenon independently explored in concurrent work proposing the Universal Weight Subspace Hypothesis~\cite{kaushik2025universal}. This hypothesis posits that many models naturally share significant similarity in their weight matrices' principal components (spectral subspaces), regardless of initialization.

Our study consisted of four distinct experimental runs, three of them utilizing a task vector derived from the fine-tuning on the Cars dataset:

\begin{itemize}
    \item[\textbf{A.}] \textbf{Baseline:} We measured the fine-tuning performance of a randomly selected ViT-B/16 Cars task vector on all target tasks, using the standard ViT-B/16 base model. This served as our internal compatibility benchmark.
    \item[\textbf{B.}] \textbf{Cross-Initialization Transfer (Minor Architectural Changes):} We tested transfer between two distinct base models (ViT-B/32 and ViT-B/16) that share the same pre-training objective (CLIP) but represent different initializations and minor architectural differences (patch size). The task vector was computed as the difference between the fine-tuned and base parameters for ViT-B/32: $\Delta_{\text{task}} = \theta_{\text{ViT-B-32, Cars}} - \theta_{\text{ViT-B-32, base}}$. We applied our full AXIS adaptation methodology to the ViT-B/16 base model, explicitly skipping the two layers with shape mismatches while transferring knowledge from all other compatible layers. This configuration simultaneously tests cross-initialization transfer and adaptability to minor architectural differences.
    \item[\textbf{C.}] \textbf{Different Pre-training (Same Source Task):} To test the necessity of a shared initialization history, we utilized an architecturally compatible model with an entirely different pre-training source. We took the OpenClip ViT-B/16 model and fine-tuned it on the Cars dataset using the standard aTLAS recipe to create a new source task vector. The task vector was calculated as $\Delta_{\text{task}} = \theta_{\text{OpenClip, Cars}} - \theta_{\text{OpenClip, base}}$, and then applied to the CLIP ViT-B/16 base model without altering AXIS procedure.
    \item[\textbf{D.}] \textbf{Out-of-Distribution Control:} As a control, we employed Microsoft's popular BiomedCLIP (ViT-B/16) from HuggingFace. This domain-specific model, fine-tuned on medical images, serves as a highly distant reference point. The model is architecturally compatible but pre-trained and fine-tuned on data completely unrelated to our broad category of target tasks. We computed the task vector as the difference between the fine-tuned and base parameters of OpenClip (similar to C) and applied it to the CLIP ViT-B/16 base model.
\end{itemize}

The results for $N$=0.1 are presented in Table~\ref{tab:cross_init}, which distinguishes the impact of the source task vector across the tested boundaries.

\begin{table}[H]
\centering
\caption{Transfer performance of the different task vectors under varying initialization and architecture constraints ($N$=0.1).}
\begin{tabular}{l c c c c}
\toprule
\textbf{Dataset} & \textbf{A. ViT-B/16} & \textbf{B. ViT-B/32} & \textbf{C. OpenCLIP ViT-B/16} & \textbf{D. BiomedCLIP} \\
\midrule
CIFAR10 & 97.17\% & 97.30\% & 97.11\% & 33.84\% \\
CIFAR100 & 80.73\% & 80.91\% & 80.37\% & 5.71\% \\
CUB200 & 63.43\% & 61.93\% & 61.13\% & 0.90\% \\
Caltech101 & 96.14\% & 95.62\% & 94.87\% & 17.45\% \\
Caltech256 & 91.49\% & 91.39\% & 90.85\% & 5.23\% \\
Country211 & 23.89\% & 24.11\% & 23.40\% & 1.09\% \\
DTD & 59.84\% & 54.15\% & 54.63\% & 4.41\% \\
EuroSAT & 97.56\% & 97.19\% & 97.19\% & 51.30\% \\
FGVCAircraft & 34.05\% & 30.42\% & 30.78\% & 1.35\% \\
Flowers102 & 78.13\% & 74.37\% & 74.09\% & 0.47\% \\
Food101 & 91.01\% & 91.16\% & 90.96\% & 3.80\% \\
GTSRB & 90.91\% & 88.33\% & 87.93\% & 14.45\% \\
MNIST & 98.33\% & 97.98\% & 97.95\% & 20.60\% \\
OxfordIIITPet & 93.40\% & 93.49\% & 93.21\% & 3.82\% \\
PascalVOC & 88.42\% & 88.17\% & 87.87\% & 36.79\% \\
RESISC45 & 89.46\% & 88.22\% & 88.65\% & 14.30\% \\
STL10 & 99.02\% & 99.12\% & 99.08\% & 27.29\% \\
SUN397 & 71.26\% & 70.02\% & 70.26\% & 0.63\% \\
SVHN & 90.56\% & 89.49\% & 90.13\% & 24.51\% \\
UCF101 & 76.92\% & 74.83\% & 72.80\% & 4.47\% \\
\midrule
\textbf{Average} & \textbf{80.59\%} & \textbf{79.41\%} & \textbf{79.16\%} & \textbf{13.62\%} \\
\bottomrule
\end{tabular}
\label{tab:cross_init}
\end{table}

Overall, the results demonstrate that AXIS successfully achieves knowledge transfer despite differences in initialization and minor architectural changes. The hierarchy of transfer compatibility is clear: the highest performance benchmark remains the most compatible version (the native ViT-B/16 baseline), followed closely by the ViT-B/32 task vector and the OpenClip task vector. This shows a consistent relationship between base model compatibility and final performance, with both cross-initialization and minor architectural changes only leading to a slight decrease in average accuracy (from $80.59\%$ to $79.41\%$ and $79.16\%$, respectively).

The performance of the domain-distant BiomedCLIP source, while low on average, confirmed its role as an extreme sanity check. Crucially, even this highly specialized model provided positive transfer, exceeding the zero-shot capabilities of the ViT-B/16 base model on certain target tasks (e.g., EuroSAT). This result, expected given the model's highly specific domain knowledge, confirms that our method could function robustly even when provided with an extremely low-relevance source model.

The decision to utilize only a single source task vector for these boundary tests was deliberate, allowing us to avoid the methodological ambiguity that incorporating many other compatible source task vectors might mask poor performance.

\section{Transferability by SVD components}
\label{sec:svd_components_analysis}

We operate under the hypothesis that for models sharing a common initialization, the magnitude of a singular value serves as a robust proxy for the information content of that component. Larger deviations from the pre-trained state indicate structural updates necessary for the source task, whereas smaller values likely correspond to task-specific noise or minor adjustments. Thus, our global ranking strategy naturally prioritizes the most significant adaptive signals across all sources.

\subsection{Individual component analysis}

In this experiment, we incrementally tested the performance of single components. Specifically, during Stage 2, we fine-tuned only one singular value per 2D layer in $\Delta_{target}$. This allowed us to evaluate which structural components contribute most effectively to the target task.

\begin{figure}[H]
    \centering
    \includegraphics[width=0.99\columnwidth]{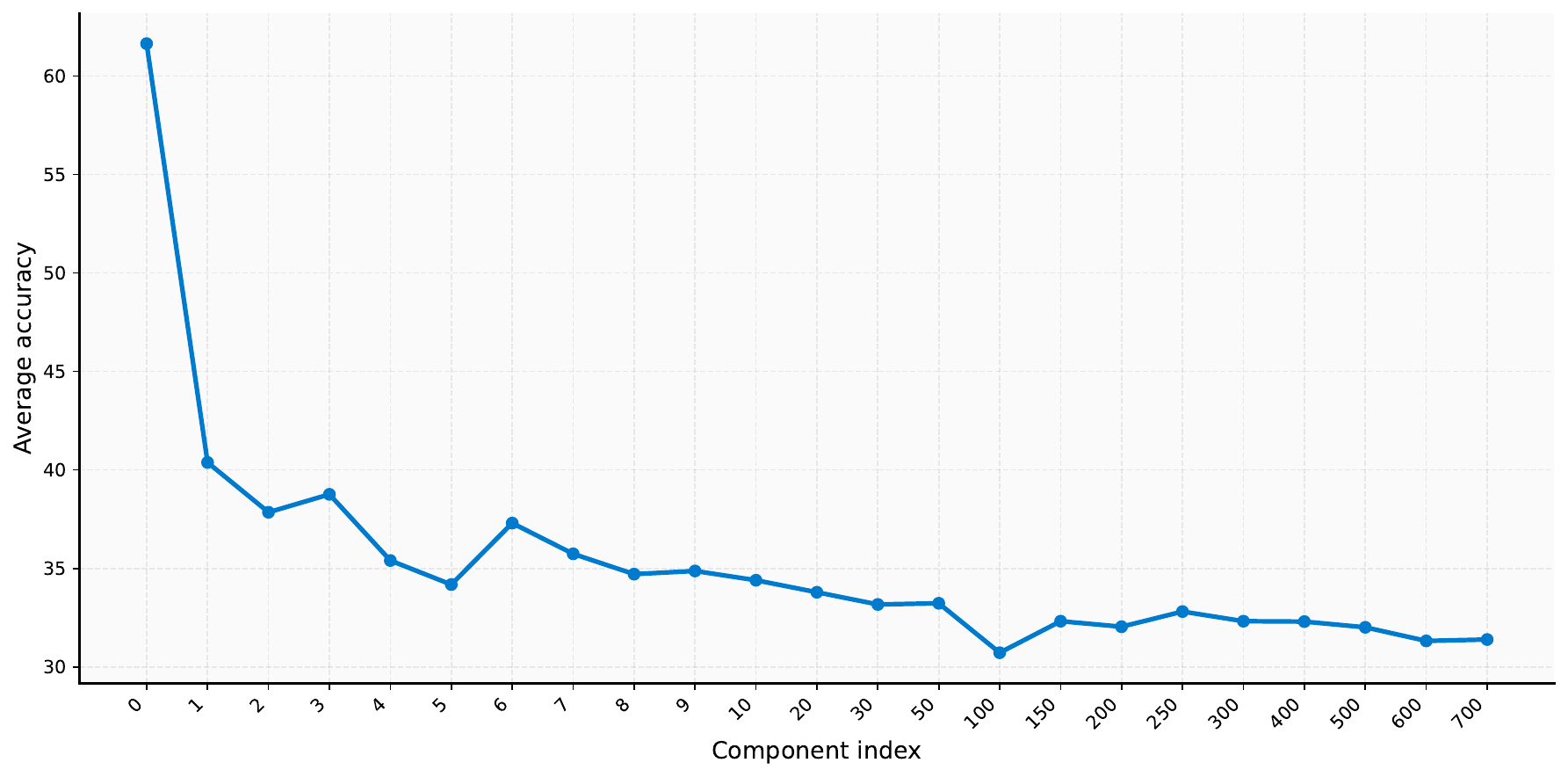}
    \caption{Impact of singular value rank on transfer performance. The plot illustrates the average accuracy obtained when fine-tuning individual singular components isolated by rank. The sharp decline demonstrates that the most structurally dominant components encapsulate the majority of transferable knowledge.}
    \label{fig:cal_top_k_exp}
\end{figure}

The averaged performance across all target tasks is illustrated in Figure~\ref{fig:cal_top_k_exp}. The results demonstrate a clear trend: higher raw singular values correlate with a significantly better capacity to transfer knowledge. As observed in the figure, the most structurally dominant component achieves an accuracy of approximately 61.6\%, whereas performance drops sharply for lower-ranked components. 

\subsection{Ground-truth alignment analysis}

We empirically demonstrate that the singular vectors associated with the largest singular values in source tasks offer the highest transferability to unseen target tasks. To quantify the transferability of specific components, we conduct an analysis where we assume access to the ground-truth target task vector, denoted as $\Delta_{target}$. This allows us to measure how well specific components from source tasks can capture the target task update. For each source task $i$, we decompose its task matrix via SVD:
\begin{equation}
\Delta_{source}^{(i)} = \sum_{k=1}^{r_{i}} \sigma_{k}^{(i)} \mathbf{u}_{k}^{(i)} (\mathbf{v}_{k}^{(i)})^\top,
\end{equation}
where the singular values $\sigma_{k}^{(i)}$ are sorted in descending order and $\mathbf{u}_{k}^{(i)},\mathbf{v}_{k}^{(i)}$ are the left and right singular vectors. We evaluate for each layer $l$ the transferability of each rank‑1 component 
using the Preserved Energy metric (independent of the component's magnitude $\sigma_{k}^{(i)}$):
\begin{equation}
E(k,i) = \frac{\langle \Delta_{target}, \mathbf{u}_{k}^{(i)} (\mathbf{v}_{k}^{(i)})^\top \rangle^2}{\| \Delta_{target} \|_F^2},
\end{equation}
To visualize the relationship between component rank $k$ and transferability, we aggregate $E(k,i)$ across all layers and tasks, ordering components by their original rank.

\begin{figure}[H]
    \centering
    \includegraphics[width=0.99\columnwidth]{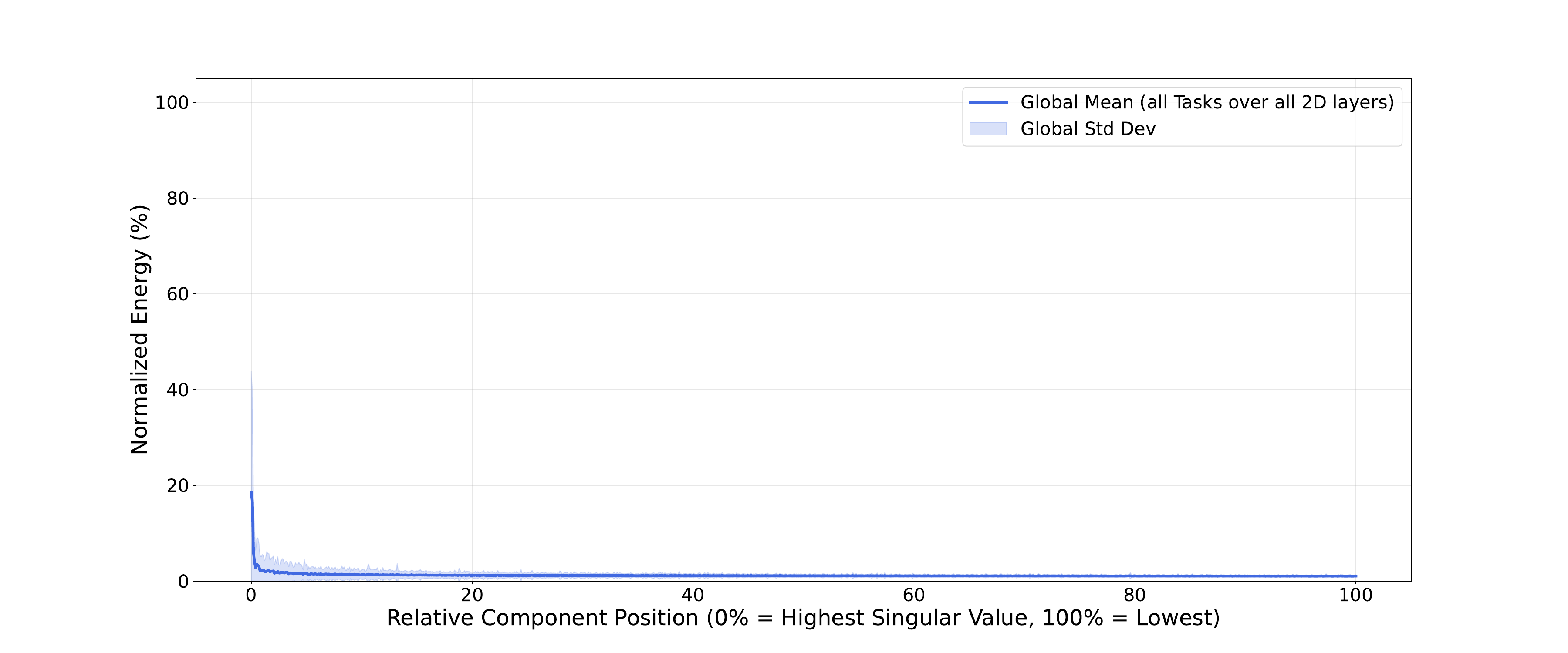}
    \caption{Normalized Energy Distribution (Preserved Energy) as a function of component rank. The x-axis represents the component index, sorted by original singular value magnitude (descending), while the y-axis shows the projection of source components onto the target update $\Delta_{target}$. The displayed values are averaged across all source models and unseen target tasks for every layer matrix. }
    \label{fig:oracle_top_k_exp}
\end{figure}

The analysis reveals a distinct concentration of high preserved energy at the lower ranks, specifically corresponding to the components with the highest singular values (i.e., $k=1, 2, \dots$) in the source tasks. This initial peak is followed by a rapid decay and subsequent plateau at higher indices. Importantly, this distribution remains consistent when averaged across all target tasks and layers, suggesting a universal property of the task vector space. These findings demonstrate a strong empirical link between singular value magnitude and cross-task transferability, confirming that the dominant components in source tasks are consistently the ones that align best with the ground-truth direction of the target task.

\section{Incremental knowledge aggregation}
\label{sec:incremental}

To evaluate the adaptability of AXIS to dynamic scenarios where source models arrive sequentially, we examine an \textit{incremental aggregation protocol}. While the default framework performs a global ranking over the singular components of the entire pool of source task matrices $\{\Delta_1, \dots, \Delta_{T-1}\}$, the incremental variant updates the merged knowledge base iteratively.

Specifically, we initialize the merged matrix $\Delta_{m}$ with the first two source task vectors. For each subsequent incoming source $\Delta_{i}$, we treat the currently accumulated matrix $\Delta_{m}^{(i-1)}$ as a consolidated representation of prior knowledge and merge it with the new source. This recursive update rule allows us to apply the aggregation mechanism defined in Stage 1 pairwise:

\begin{equation}
    \Delta_{m}^{(i)} = \operatorname{Stage1}(\{\Delta_{m}^{(i-1)}, \Delta_{i}\})
\end{equation}

In this setup, the aggregation step selects the top-$K$ components from the union of the accumulated basis and the new task vector. Crucially, this creates a memory-efficient online process where historical source parameters are discarded immediately after integration, eliminating the need for a persistent buffer of previous models.

With an average accuracy of 78.55\%, the streaming approach performs on par with the standard global ranking protocol (78.48\%). Consequently, AXIS demonstrates effectiveness at real-time structural knowledge accumulation, eliminating the overhead associated with iteratively processing the full model history.

\section{In-depth robustness analyses}
\label{sec:robustness}
\subsection{Robustness to Input Perturbations} 
\begin{figure}[H]
    \centering
    \includegraphics[width=0.60\textwidth]{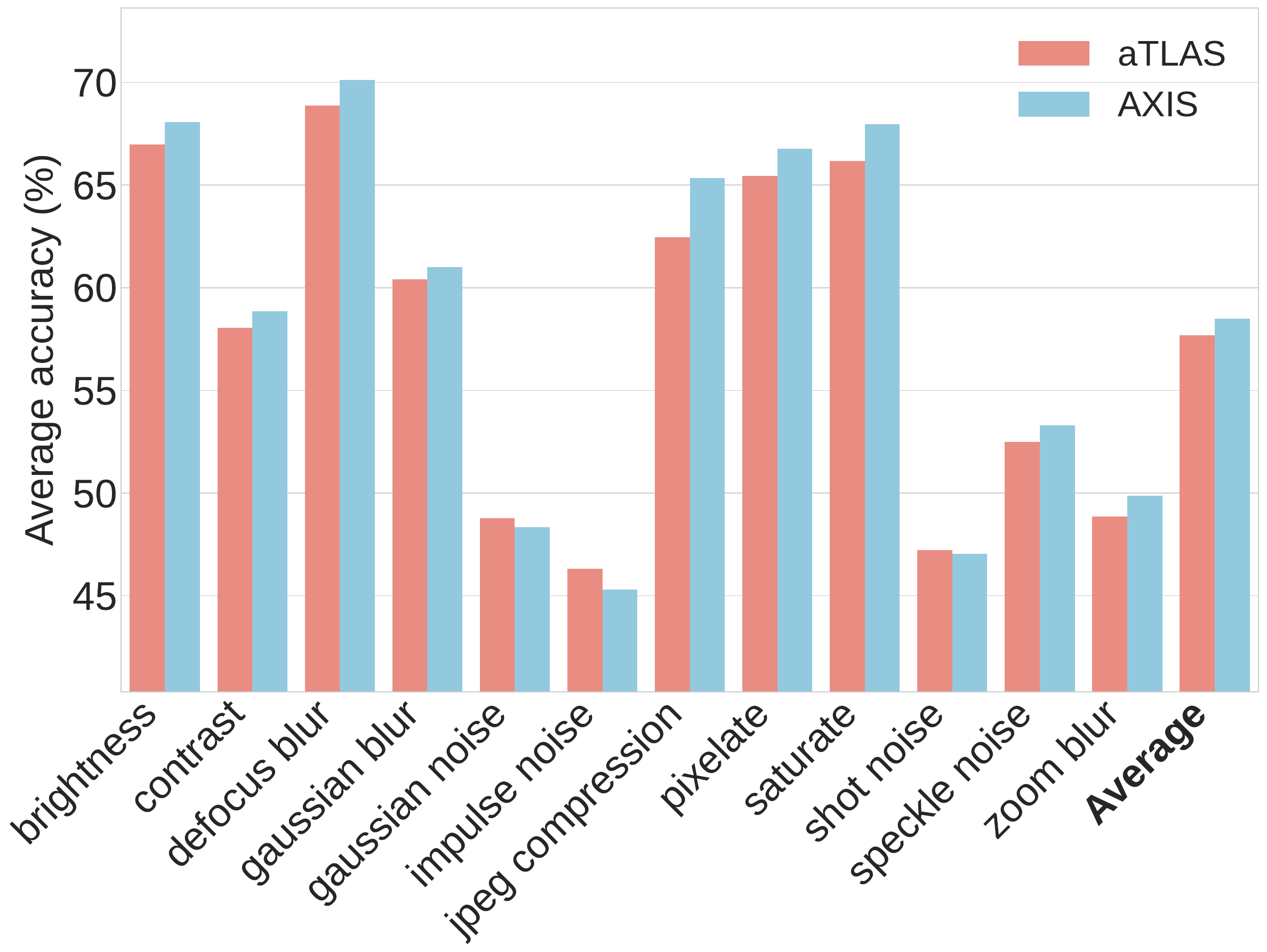}
    \caption{The accuracy across each type of corruption is evaluated for all severity levels ranging from 1 to 5 for all 21 target tasks.}
    \label{fig:image_corruptions}
\end{figure}

\begin{figure}[H]
    \centering
    \begin{minipage}[t]{0.48\textwidth}
        \centering
        \includegraphics[width=\linewidth]{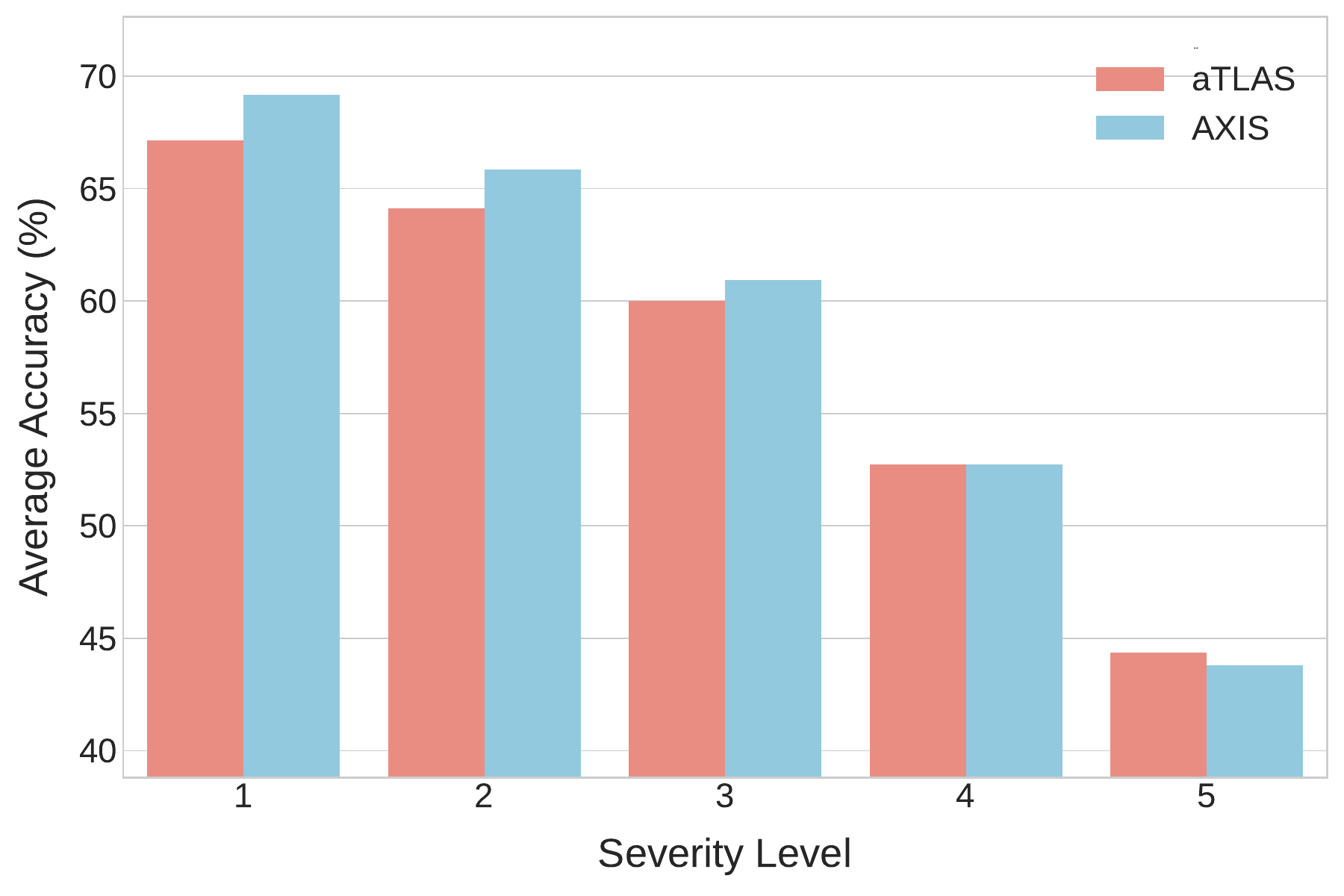}
        \caption{Severity levels average over all 12 image corruptions.}
        \label{fig:severity_levels}
    \end{minipage}
    \hfill 
    \begin{minipage}[t]{0.48\textwidth}
        \centering
        \includegraphics[width=\linewidth]{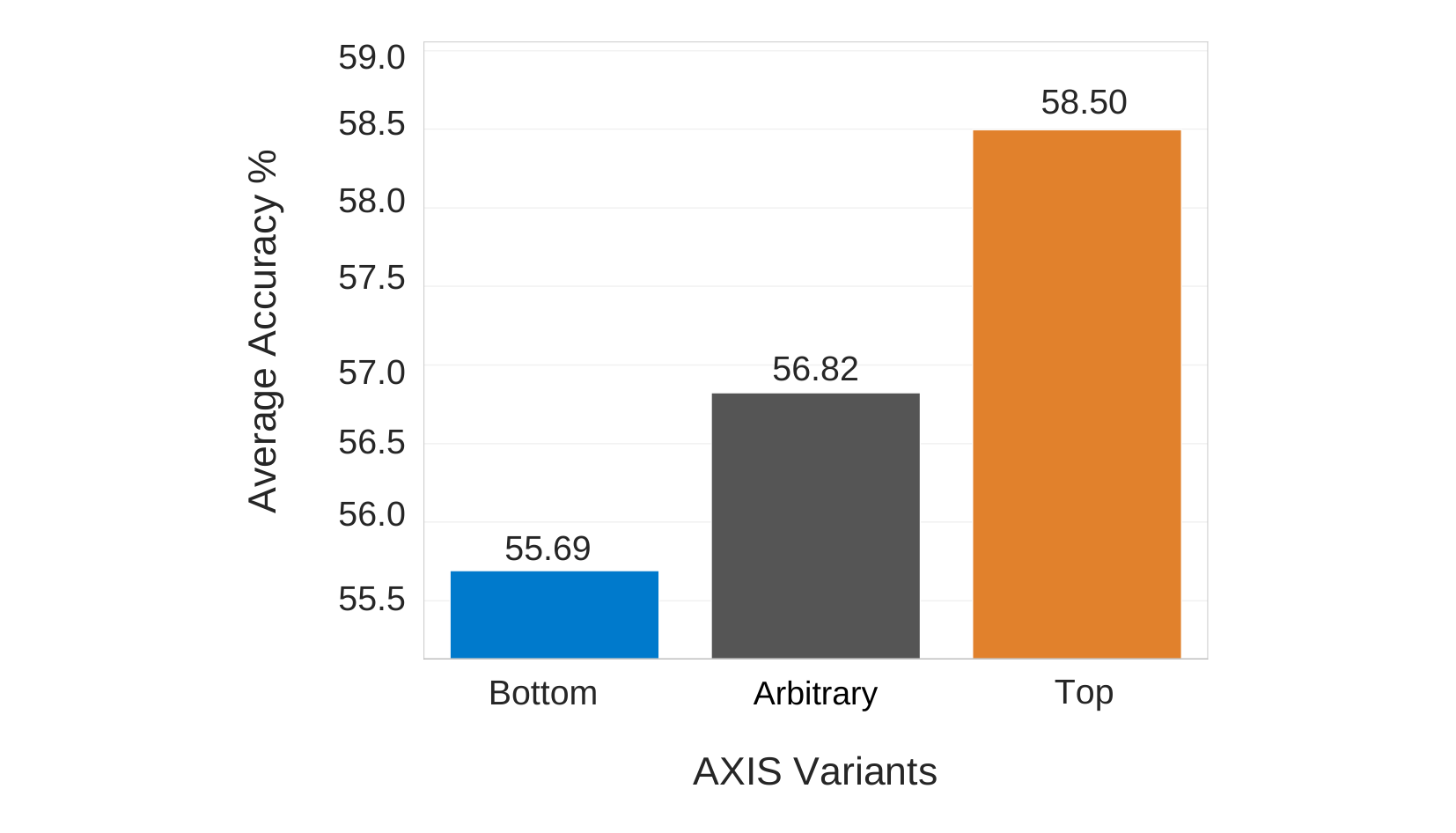}
        \caption{AXIS with top-component selection is more robust against common corruptions than other AXIS's aggregation strategies.}
        \label{fig:top_axis_common_corruption}
    \end{minipage}
\end{figure}

To further probe the robustness capabilities of AXIS and aTLAS, we evaluate them against a set of 12 common image corruptions~\cite{hendrycks2019benchmarking}. Each corruption type is applied to the test set of target task images at five distinct severity levels to simulate a range of degradations. As illustrated in Figure~\ref{fig:image_corruptions}, our proposed method, AXIS, maintains a slight average performance advantage (0.83 percentage points). This margin is particularly pronounced for corruption types where overall accuracy remains high, indicating better robustness in moderately challenging conditions. A detailed breakdown by severity level delineates this trend more clearly (see Figure~\ref{fig:severity_levels}). AXIS demonstrates greater resilience across the initial four perturbation levels, outperforming aTLAS by margins of 2.04 percentage points for the lowest corruption severity.

Furthermore, we extend our robustness evaluation to scenarios with partial input information, a challenge simulated using patch dropout. A detailed, step-by-step analysis, presented in Table~\ref{tbl:appendix_robustness_detailed}, illustrates how the model's resilience to input masking evolves as the incremental aggregation of each source task vector is performed. This granular breakdown demonstrates that the fusion of diverse knowledge sources enhances the model's ability to perform predictions even when significant portions of the input are omitted.

\begin{table*}[h]
    \centering
    \caption{Performance analysis of AXIS under increasing input masking. The table illustrates that aggregating more source task vectors (TV) enhances model robustness to input patch dropout. We report the mean accuracy (\%) across all target tasks for dropout rates from 0\% to 50\%. Each row corresponds to a different number of aggregated sources, and values in parentheses show the improvement in percentage points (p.p.) over the first, single task vector baseline (first row).}
    \scriptsize
    \setlength{\tabcolsep}{2.5pt}
    \begin{adjustbox}{width=\textwidth}
        \begin{tabular}{l c c c c c c c c c c}
            \toprule
            \textbf{TV} & \multicolumn{10}{c}{\textbf{Input Patch Dropout (\%)}} \\
            \cmidrule(l){2-11}
        1 & 77.31 & 75.53 & 72.78 & 67.51 & 63.61 & 56.86 & 52.11 & 44.66 & 39.79 & 28.74 \\
        2 & 77.51 \small(+0.20) & 75.80 \small(+0.27) & 73.40 \small(+0.62) & 69.04 \small(+1.53) & 65.38 \small(+1.77) & 59.54 \small(+2.68) & 55.17 \small(+3.05) & 48.36 \small(+3.69) & 43.63 \small(+3.84) & 32.67 \small(+3.93) \\
        3 & 77.52 \small(+0.21) & 75.88 \small(+0.35) & 73.49 \small(+0.71) & 69.27 \small(+1.76) & 65.72 \small(+2.12) & 60.00 \small(+3.13) & 55.87 \small(+3.75) & 49.28 \small(+4.62) & 44.84 \small(+5.05) & 34.08 \small(+5.34) \\
        4 & 77.85 \small(+0.54) & 76.12 \small(+0.60) & 73.68 \small(+0.89) & 69.43 \small(+1.91) & 65.92 \small(+2.31) & 59.97 \small(+3.10) & 55.74 \small(+3.62) & 49.12 \small(+4.45) & 44.58 \small(+4.79) & 33.49 \small(+4.75) \\
        5 & 77.81 \small(+0.50) & 76.24 \small(+0.71) & 73.98 \small(+1.20) & 69.82 \small(+2.30) & 66.51 \small(+2.90) & 60.95 \small(+4.08) & 56.87 \small(+4.76) & 50.27 \small(+5.60) & 45.65 \small(+5.86) & 34.59 \small(+5.86) \\
        6 & 78.02 \small(+0.71) & 76.41 \small(+0.88) & 73.96 \small(+1.18) & 69.78 \small(+2.27) & 66.40 \small(+2.80) & 60.53 \small(+3.66) & 56.54 \small(+4.42) & 50.10 \small(+5.43) & 45.63 \small(+5.84) & 34.64 \small(+5.90) \\
        7 & 78.08 \small(+0.76) & 76.48 \small(+0.95) & 74.23 \small(+1.45) & 70.00 \small(+2.49) & 66.82 \small(+3.21) & 61.16 \small(+4.30) & 57.01 \small(+4.90) & 50.43 \small(+5.77) & 45.74 \small(+5.95) & 34.45 \small(+5.71) \\
        8 & 78.33 \small(+1.02) & 76.67 \small(+1.14) & 74.27 \small(+1.49) & 70.27 \small(+2.76) & 66.96 \small(+3.35) & 61.21 \small(+4.35) & 57.13 \small(+5.01) & 50.71 \small(+6.04) & 46.12 \small(+6.33) & 34.98 \small(+6.24) \\
        9 & 78.41 \small(+1.10) & 76.74 \small(+1.21) & 74.42 \small(+1.64) & 70.29 \small(+2.78) & 66.88 \small(+3.27) & 61.49 \small(+4.63) & 57.63 \small(+5.52) & 51.24 \small(+6.57) & 47.02 \small(+7.23) & 36.10 \small(+7.36) \\
        10 & 78.16 \small(+0.85) & 76.60 \small(+1.07) & 74.20 \small(+1.42) & 69.85 \small(+2.33) & 66.37 \small(+2.77) & 60.55 \small(+3.69) & 56.32 \small(+4.20) & 49.77 \small(+5.11) & 45.17 \small(+5.38) & 34.25 \small(+5.51) \\
        11 & 78.40 \small(+1.09) & 76.87 \small(+1.34) & 74.44 \small(+1.66) & 70.29 \small(+2.78) & 66.81 \small(+3.20) & 60.74 \small(+3.88) & 56.21 \small(+4.09) & 49.26 \small(+4.59) & 44.47 \small(+4.68) & 32.81 \small(+4.07) \\
        12 & 78.51 \small(+1.20) & 76.90 \small(+1.38) & 74.56 \small(+1.78) & 70.34 \small(+2.83) & 66.92 \small(+3.32) & 61.11 \small(+4.24) & 57.05 \small(+4.93) & 50.31 \small(+5.65) & 45.78 \small(+5.99) & 33.99 \small(+5.26) \\
        13 & 78.37 \small(+1.06) & 76.71 \small(+1.19) & 74.34 \small(+1.55) & 70.15 \small(+2.63) & 66.78 \small(+3.17) & 61.02 \small(+4.15) & 56.85 \small(+4.74) & 49.97 \small(+5.31) & 45.32 \small(+5.53) & 33.81 \small(+5.08) \\
        14 & 78.41 \small(+1.10) & 76.83 \small(+1.30) & 74.42 \small(+1.64) & 70.16 \small(+2.65) & 66.75 \small(+3.15) & 60.87 \small(+4.00) & 56.71 \small(+4.60) & 49.65 \small(+4.98) & 44.90 \small(+5.10) & 33.05 \small(+4.31) \\
        15 & 78.34 \small(+1.02) & 76.81 \small(+1.28) & 74.50 \small(+1.71) & 70.28 \small(+2.77) & 66.82 \small(+3.21) & 60.74 \small(+3.87) & 56.32 \small(+4.20) & 49.24 \small(+4.57) & 44.64 \small(+4.85) & 33.14 \small(+4.40) \\
        16 & 78.42 \small(+1.11) & 76.85 \small(+1.32) & 74.70 \small(+1.92) & 70.45 \small(+2.94) & 67.11 \small(+3.50) & 61.37 \small(+4.50) & 56.96 \small(+4.85) & 50.16 \small(+5.50) & 45.73 \small(+5.94) & 34.51 \small(+5.78) \\
        17 & 78.41 \small(+1.09) & 76.82 \small(+1.29) & 74.57 \small(+1.79) & 70.38 \small(+2.87) & 67.06 \small(+3.45) & 61.32 \small(+4.45) & 57.10 \small(+4.98) & 50.41 \small(+5.74) & 45.93 \small(+6.14) & 34.91 \small(+6.17) \\
        18 & 78.54 \small(+1.23) & 76.94 \small(+1.41) & 74.63 \small(+1.85) & 70.53 \small(+3.01) & 67.36 \small(+3.76) & 61.77 \small(+4.91) & 57.62 \small(+5.51) & 50.92 \small(+6.26) & 46.45 \small(+6.66) & 34.92 \small(+6.19) \\
        19 & 78.58 \small(+1.27) & 76.91 \small(+1.38) & 74.61 \small(+1.83) & 70.20 \small(+2.69) & 66.87 \small(+3.26) & 61.19 \small(+4.32) & 56.97 \small(+4.86) & 50.48 \small(+5.82) & 46.05 \small(+6.26) & 34.64 \small(+5.90) \\
        20 & 78.50 \small(+1.19) & 76.75 \small(+1.22) & 74.51 \small(+1.73) & 70.14 \small(+2.63) & 66.93 \small(+3.32) & 61.19 \small(+4.33) & 57.25 \small(+5.13) & 50.58 \small(+5.91) & 46.31 \small(+6.52) & 35.05 \small(+6.31) \\
        \textbf{21} & \textbf{78.48} \small(+1.16) & \textbf{76.82} \small(+1.29) & \textbf{74.68} \small(+1.90) & \textbf{70.52} \small(+3.00) & \textbf{67.37} \small(+3.76) & \textbf{61.84} \small(+4.97) & \textbf{57.82} \small(+5.71) & \textbf{51.18} \small(+6.51) & \textbf{46.63} \small(+6.83) & \textbf{35.39} \small(+6.65) \\
        \bottomrule
    \end{tabular}
    \end{adjustbox}
    \label{tbl:appendix_robustness_detailed}
\end{table*}

\subsection{Training Data Availability}

\begin{figure}[H]
    \centering
    \includegraphics[width=0.49\textwidth]{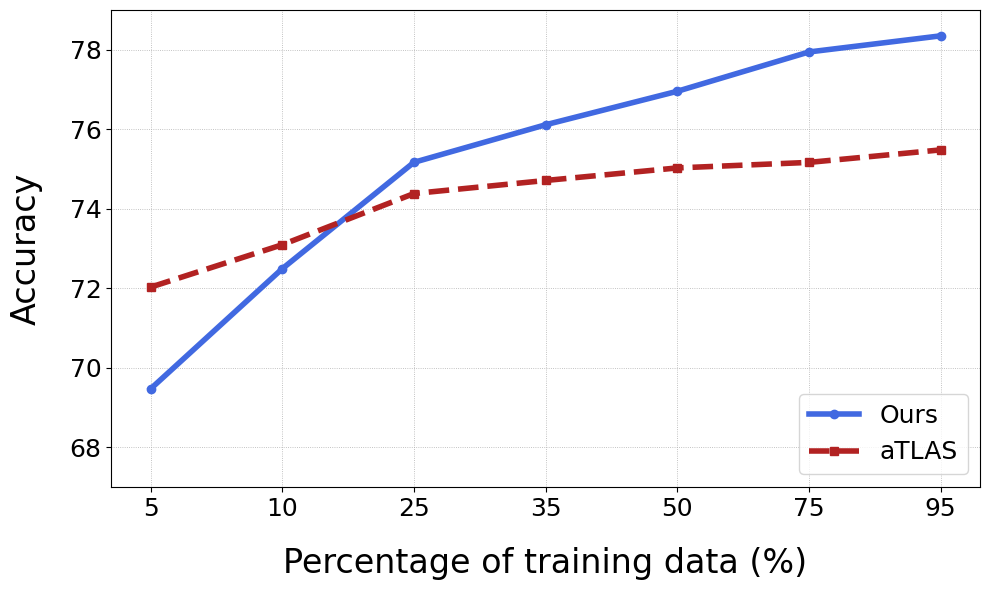} 
    \caption{AXIS performs better with smaller amounts of training data in almost all cases.}
    \label{fig:data_availability}
\end{figure}

To assess the data efficiency of our approach and its robustness in limited data scenarios, we investigate the performance of our method compared to aTLAS under varying levels of training data availability for the target task. For this experiment, we reduce the size of the target task's training dataset, creating subsets with 5\%, 10\%, 25\%, 50\%, 75\%, and 95\% of the original samples. The results, illustrated in Figure~\ref{fig:data_availability}, demonstrate that our method maintains a significant performance advantage over aTLAS across the broad majority of data availability levels.


\subsection{Robustness Against Altered Source Parameters}

\begin{figure}[H]
    \centering
    \includegraphics[width=0.7\textwidth]{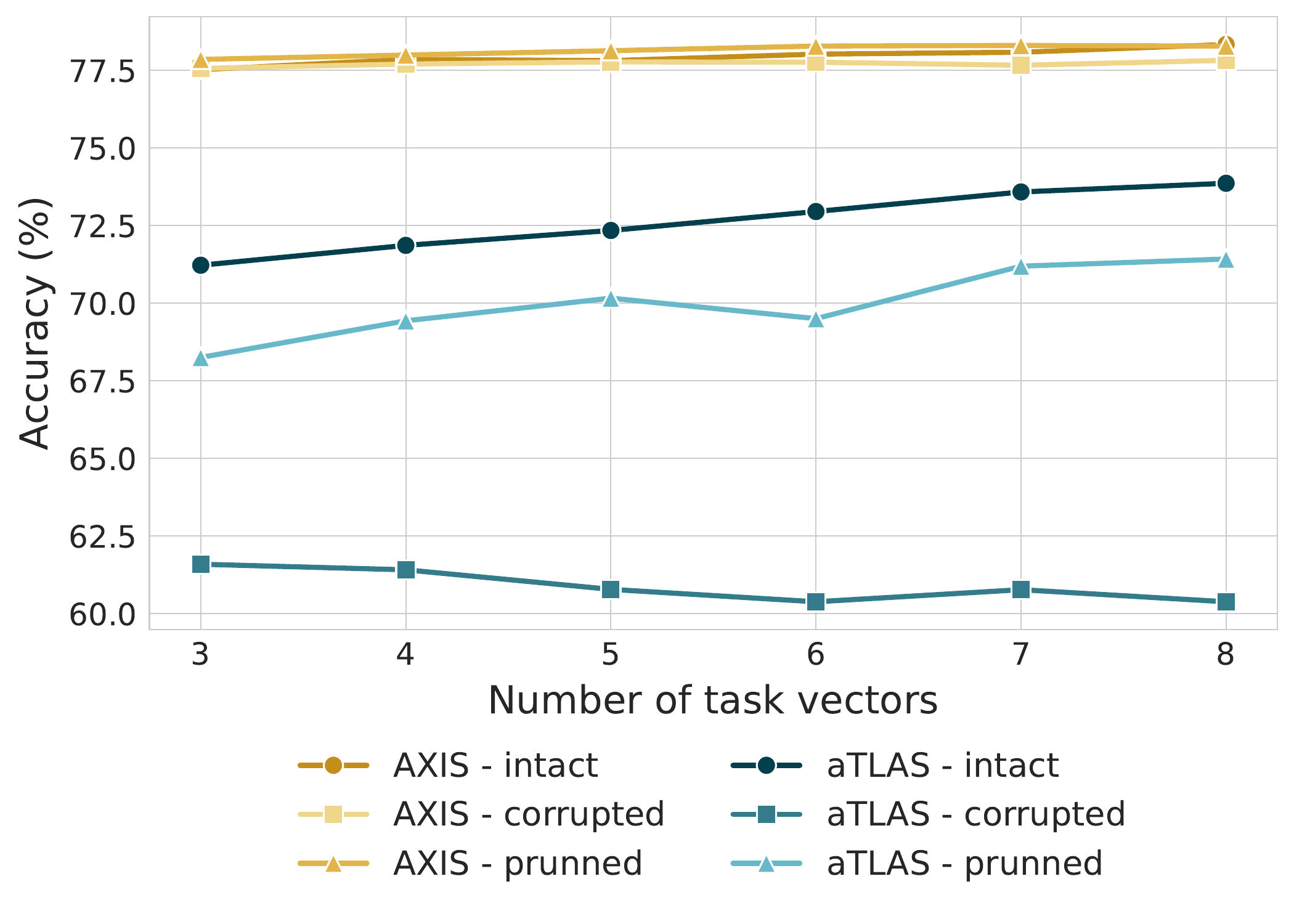}
    \caption{Robustness to altered source task vectors. The plot compares performance under two distinct perturbation scenarios, with results averaged across all 21 target tasks. Our method AXIS demonstrates substantially higher resilience to both scenarios compared to aTLAS.}
    \label{fig:corruption_robustness}
\end{figure}

For a detailed analysis of the framework's robustness, we refer to Table~\ref{tbl:robustness_comparison_adjusted} and Figure~\ref{fig:corruption_robustness}, which provides a comprehensive performance breakdown under two challenging scenarios: contamination by a single noisy source vector and aggregation of heavily pruned (95\%) source vectors.

\begin{table}[h]
    \centering
    \caption{Robustness to altered source task vectors. AXIS shows higher resilience to corruption and pruning compared to aTLAS.}
    \begin{adjustbox}{width=0.49\textwidth}
    \begin{tabular}{@{}lcccccc@{}}
    \toprule
    \multirow{2}{*}{\shortstack{Task \\ Vectors}} & \multicolumn{3}{c}{aTLAS} & \multicolumn{3}{c}{AXIS (ours)} \\
    \cmidrule(lr){2-4} \cmidrule(lr){5-7}
     & intact & corrupted & pruned & intact & corrupted & pruned \\
    \midrule
    3 & 71.22 & 61.59 & 68.25 & 77.52 & 77.56 & 77.85 \\
    4 & 71.86 & 61.41 & 69.43 & 77.85 & 77.70 & 77.99 \\
    5 & 72.34 & 60.78 & 70.16 & 77.81 & 77.77 & 78.13 \\
    6 & 72.95 & 60.38 & 69.50 & 78.02 & 77.76 & 78.28 \\
    7 & 73.58 & 60.77 & 71.19 & 78.08 & 77.66 & 78.30 \\
    8 & 73.86 & 60.38 & 71.42 & 78.33 & 77.82 & 78.28 \\
    \bottomrule
    \end{tabular}
    \end{adjustbox}
    \label{tbl:robustness_comparison_adjusted}
\end{table}

\section{Component selection}
\label{sec:component_selection}

We evaluated the impact of different component selection and aggregation strategies on final model performance. The goal was to ensure that our default approach, aggregating components with the highest singular values, is competitive with other plausible alternatives, especially with the highest number of source task vectors. We compared the following seven strategies:

\begin{itemize}
    \item \textbf{Top Components (our default):} As described in the main paper, we perform a global ranking of all singular components from all source tasks and select the top-K based on their singular values $(\sigma_k)$ to form the merged matrix $\Delta_m = \sum_{k=1}^{K} u_k \sigma_k v_k^{\top}$.

    \item \textbf{Bottom Components:} A control strategy where we select the K components with the lowest singular values from the global ranking.

    \item \textbf{Arbitrary Components:} A second control strategy where K components are arbitrarily selected from the global pool.

    \item \textbf{Average Top Components:} This baseline first distills each source task matrix $\Delta_i$ into its top-K principal components. Next, all these resulting low-rank matrices are averaged into a single matrix. Finally, we perform a new SVD on this averaged matrix and select its top-K components to form the final $\Delta_m$.

    \item \textbf{Average Bottom Components:} The inverse of the "average top components" baseline, used as a control. First, each source task matrix is reduced to a low-rank approximation using only its own bottom-K singular components. Second, these resulting low-rank matrices are averaged, and a final selection of the bottom-K components is performed via SVD on this single, averaged matrix.

    \item \textbf{Equal Top Contribution:} This strategy ensures a balanced representation from all source tasks. Instead of a global ranking, it selects an equal number of the top singular components from each individual source task. If the total budget is K components and there are $T-1$ sources, we select the top $K/(T-1)$ components from each task. These are then pooled and summed to form $\Delta_m$.
    
\end{itemize}

\begin{figure}[h!]
    \centering
    \includegraphics[width=0.85\textwidth]{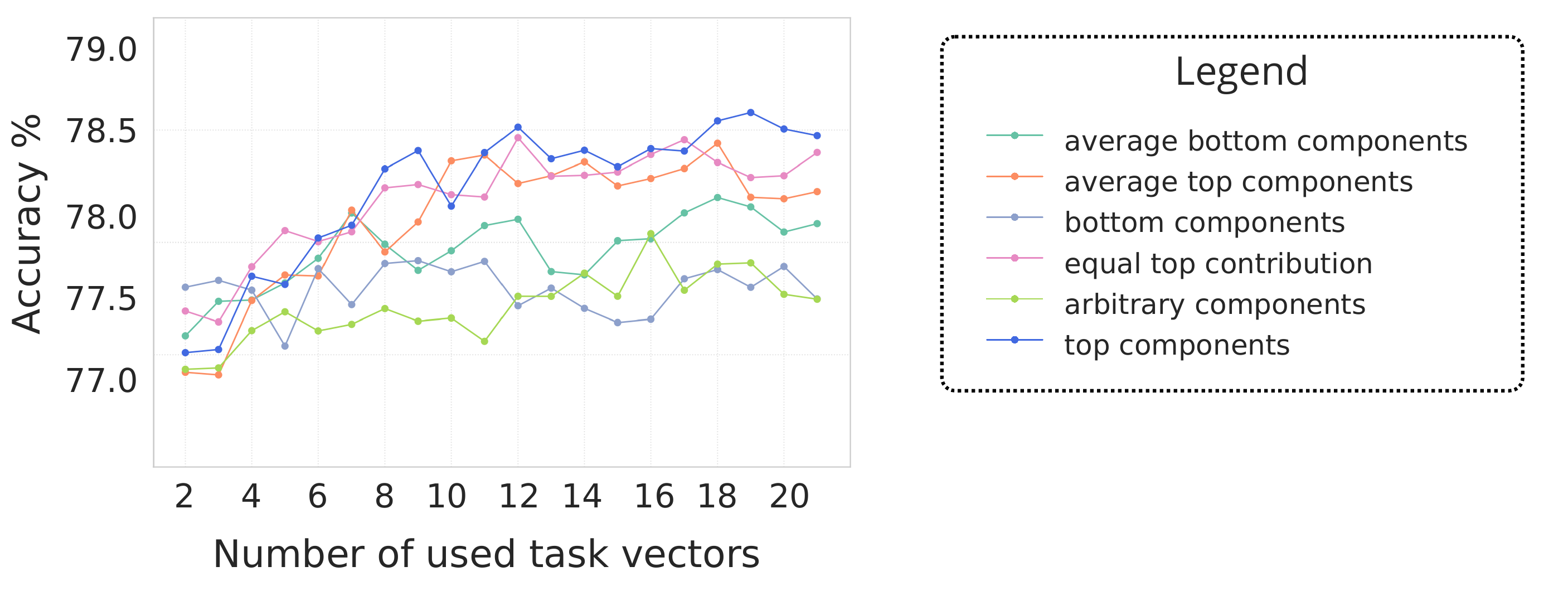}
    \caption{Performance comparison of six different SVD component aggregation strategies with $N$=10\%. The plot shows the average accuracy across all target tasks as the number of used source task vectors increases. Our default strategy, top components, yields the best performance with the largest number of sources.}
    \label{fig:all_selection_strategies}
\end{figure}

\begin{figure}[H]
    \centering
    \includegraphics[width=0.6\columnwidth]{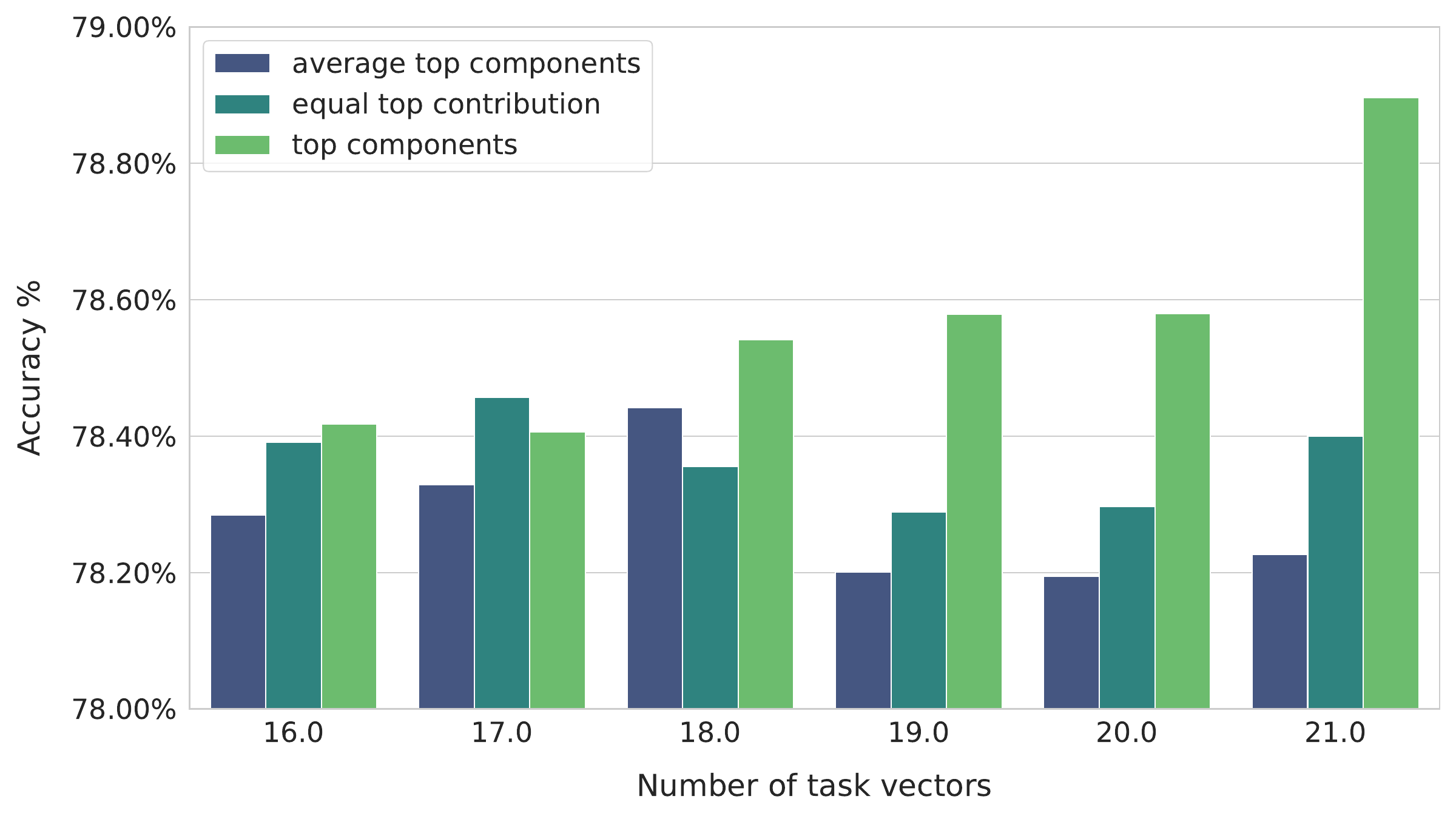}
    \caption{Detailed performance comparison of SVD component aggregation strategies, focusing on small variations within top components. While all strategies show comparable performance, the top components generally maintain a slight edge, particularly with a higher number of aggregated source tasks.}
    \label{fig:components_selection}
\end{figure}

Additionally, we compare how different selection strategies for the top-ranking components affect accuracy when using the largest number of source task vectors, as illustrated in Figure~\ref{fig:components_selection}. For this configuration, the top components strategy yielded the highest accuracy. These results are averaged across all target tasks. Additionally, we provided detailed results on the main aggregation strategies per target dataset in the Table~\ref{tab:appendix_full_results_comparison}.

\section{Detailed main results}
\label{sec:detailed_results}

\begin{table}[H]
\centering
\caption{The table clarifies that aTLAS holds a marginal advantage on only 3 datasets (OxfordIIITPet, PascalVOC, STL10), with two of these (PascalVOC, STL10) being statistically negligible, likely falling within the variance of a single-seed run. AXIS's superior per-task performance with ViT-B-32 architecture is visually detailed in the spider plot on the right side of Figure~\ref{fig:vit-b-32_comparison}.}
\begin{tabular}{lccc}
\toprule
\textbf{Dataset} & \textbf{AXIS (\%)} & \textbf{aTLAS (\%)} & \textbf{Absolute gain (AXIS)} \\
\midrule
CIFAR100 & 80.13 & 79.14 & +0.99 \\
CIFAR10 & 96.78 & 96.51 & +0.27 \\
CUB200 & 59.06 & 55.82 & +3.24 \\
Caltech101 & 94.47 & 94.18 & +0.29 \\
Caltech256 & 88.88 & 88.09 & +0.79 \\
Cars & 65.30 & 62.82 & +2.48 \\
Country211 & 18.28 & 18.08 & +0.20 \\
DTD & 68.09 & 55.69 & +12.40 \\
EuroSAT & 97.74 & 95.67 & +2.07 \\
FGVCAircraft & 30.63 & 24.96 & +5.67 \\
Flowers102 & 78.37 & 70.24 & +8.13 \\
Food101 & 86.23 & 85.63 & +0.60 \\
GTSRB & 90.58 & 76.96 & +13.62 \\
MNIST & 98.03 & 95.30 & +2.73 \\
RESISC45 & 87.76 & 78.95 & +8.81 \\
SUN397 & 67.62 & 66.50 & +1.12 \\
SVHN & 88.65 & 86.32 & +2.33 \\
UCF101 & 74.31 & 69.60 & +4.71 \\
OxfordIIITPet & 90.79 & 91.82 & -1.03 \\
PascalVOC & 87.16 & 87.18 & -0.02 \\
STL10 & 97.91 & 98.33 & -0.42 \\

\midrule
\textbf{Average} & \textbf{78.42} & \textbf{75.13} & \textbf{+3.29} \\
\bottomrule
\end{tabular}
\label{tab:performance_comparison_16tv}
\end{table}

For a comprehensive and granular evaluation of our proposed framework, Tables~\ref{tab:results_part1}--\ref{tab:results_part3} present a detailed, per-dataset comparison of AXIS and the aTLAS baseline. 

\begin{table*}[!htbp]
\centering
\caption{Detailed results per target dataset for various numbers of source task vectors (STV). Part 1 of 3.}
\begin{adjustbox}{width=0.8\textwidth,center}
\begin{tabular}{@{}clccccccc@{}}
\toprule
\textbf{STV} & \textbf{Method} & \textbf{CIFAR100} & \textbf{CIFAR10} & \textbf{CUB200} & \textbf{Caltech101} & \textbf{Caltech256} & \textbf{Cars} & \textbf{Country211} \\
\midrule
\multirow{6}{*}{1} & aTLAS (N=10\%) & 72.95 & 93.76 & 54.47 & 89.86 & 85.10 & 61.21 & 17.69 \\
& aTLAS (N=20\%) & 73.62 & 94.15 & 55.38 & 91.65 & 85.53 & 62.12 & 17.92 \\
& aTLAS (N=40\%) & 75.09 & 95.20 & 56.80 & 93.38 & 87.59 & 63.77 & 18.05 \\
& AXIS (N=10\%) & 77.00 & 95.85 & 57.61 & 93.89 & 88.44 & 63.54 & 17.70 \\
& AXIS (N=20\%) & 79.28 & 96.63 & 60.15 & 94.41 & 89.19 & 65.58 & 18.39 \\
& AXIS (N=40\%) & 81.45 & 97.10 & 62.50 & 94.99 & 89.38 & 65.94 & 18.64 \\
\midrule
\multirow{6}{*}{5} & aTLAS (N=10\%) & 73.90 & 94.52 & 54.83 & 91.53 & 85.43 & 62.06 & 17.78 \\
& aTLAS (N=20\%) & 74.77 & 95.17 & 55.94 & 92.68 & 87.59 & 62.53 & 18.02 \\
& aTLAS (N=40\%) & 75.29 & 95.31 & 56.85 & 93.78 & 88.06 & 63.89 & 18.17 \\
& AXIS (N=10\%) & 77.51 & 96.50 & 58.41 & 93.61 & 88.01 & 63.95 & 18.17 \\
& AXIS (N=20\%) & 79.96 & 96.84 & 59.22 & 94.70 & 89.48 & 67.23 & 18.60 \\
& AXIS (N=40\%) & 82.28 & 97.13 & 62.46 & 94.24 & 89.89 & 69.69 & 18.86 \\
\midrule
\multirow{6}{*}{10} & aTLAS (N=10\%) & 78.92 & 96.40 & 55.11 & 91.88 & 86.21 & 62.37 & 18.06 \\
& aTLAS (N=20\%) & 79.68 & 96.58 & 55.78 & 93.72 & 86.82 & 62.90 & 18.24 \\
& aTLAS (N=40\%) & 80.65 & 96.90 & 55.47 & 94.82 & 88.29 & 64.15 & 18.41 \\
& AXIS (N=10\%) & 80.09 & 96.96 & 57.85 & 94.82 & 88.76 & 64.66 & 18.24 \\
& AXIS (N=20\%) & 81.31 & 97.10 & 59.58 & 94.82 & 89.53 & 67.07 & 18.08 \\
& AXIS (N=40\%) & 82.64 & 97.49 & 61.74 & 94.64 & 89.20 & 69.92 & 19.22 \\
\midrule
\multirow{6}{*}{15} & aTLAS (N=10\%) & 78.95 & 96.46 & 55.89 & 94.70 & 88.11 & 62.04 & 18.16 \\
& aTLAS (N=20\%) & 79.81 & 96.81 & 57.08 & 95.22 & 89.19 & 64.15 & 18.30 \\
& aTLAS (N=40\%) & 80.62 & 97.19 & 57.82 & 96.08 & 89.38 & 64.88 & 18.51 \\
& AXIS (N=10\%) & 80.14 & 96.85 & 58.68 & 94.64 & 88.65 & 65.43 & 18.31 \\
& AXIS (N=20\%) & 81.55 & 97.25 & 60.94 & 95.56 & 89.89 & 66.86 & 18.48 \\
& AXIS (N=40\%) & 82.83 & 97.38 & 63.00 & 95.28 & 90.33 & 69.99 & 19.24 \\
\midrule
\multirow{6}{*}{21} & aTLAS (N=10\%) & 78.91 & 96.53 & 55.85 & 94.53 & 88.81 & 63.29 & 18.07 \\
& aTLAS (N=20\%) & 79.94 & 96.79 & 57.46 & 94.64 & 89.43 & 64.21 & 18.36 \\
& aTLAS (N=40\%) & 80.84 & 97.14 & 58.01 & 95.28 & 89.89 & 65.09 & 18.32 \\
& AXIS (N=10\%) & 80.11 & 96.93 & 58.46 & 94.99 & 88.76 & 65.09 & 18.48 \\
& AXIS (N=20\%) & 81.69 & 97.13 & 61.10 & 94.64 & 89.95 & 66.88 & 18.58 \\
& AXIS (N=40\%) & 82.96 & 97.39 & 62.70 & 95.45 & 90.75 & 70.77 & 19.42 \\
\bottomrule
\end{tabular}
\end{adjustbox}
\label{tab:results_part1}
\end{table*}

\begin{table*}[!htbp]
\centering
\caption{Detailed results per target dataset for various numbers of source task vectors (STV). Part 2 of 3.}
\begin{adjustbox}{width=0.8\textwidth,center}
\begin{tabular}{@{}clccccccc@{}}
\toprule
\textbf{STV} & \textbf{Method} & \textbf{DTD} & \textbf{EuroSAT} & \textbf{FGVCAircraft} & \textbf{Flowers102} & \textbf{Food101} & \textbf{GTSRB} & \textbf{MNIST} \\
\midrule
\multirow{6}{*}{1} & aTLAS (N=10\%) & 48.78 & 88.81 & 22.62 & 67.39 & 85.11 & 54.90 & 82.44 \\
& aTLAS (N=20\%) & 51.49 & 90.85 & 23.64 & 67.96 & 85.09 & 59.20 & 84.84 \\
& aTLAS (N=40\%) & 56.97 & 95.04 & 24.75 & 70.25 & 85.73 & 78.45 & 93.38 \\
& AXIS (N=10\%) & 67.02 & 97.30 & 29.70 & 77.49 & 85.81 & 89.57 & 97.36 \\
& AXIS (N=20\%) & 70.80 & 97.70 & 30.66 & 81.15 & 86.28 & 93.20 & 98.46 \\
& AXIS (N=40\%) & 74.15 & 98.30 & 19.65 & 81.20 & 86.93 & 94.22 & 98.76 \\
\midrule
\multirow{6}{*}{5} & aTLAS (N=10\%) & 53.03 & 94.11 & 22.86 & 68.56 & 85.27 & 66.85 & 89.08 \\
& aTLAS (N=20\%) & 54.04 & 94.48 & 24.15 & 68.26 & 85.41 & 71.35 & 91.97 \\
& aTLAS (N=40\%) & 58.67 & 95.44 & 24.83 & 69.58 & 85.86 & 79.96 & 93.44 \\
& AXIS (N=10\%) & 65.69 & 97.41 & 30.48 & 77.22 & 86.05 & 90.74 & 97.78 \\
& AXIS (N=20\%) & 70.96 & 97.63 & 33.75 & 80.09 & 86.62 & 93.45 & 98.57 \\
& AXIS (N=40\%) & 73.09 & 98.22 & 16.83 & 82.84 & 87.02 & 94.51 & 98.81 \\
\midrule
\multirow{6}{*}{10} & aTLAS (N=10\%) & 55.96 & 95.59 & 24.18 & 69.02 & 85.27 & 77.00 & 95.42 \\
& aTLAS (N=20\%) & 59.57 & 95.93 & 24.54 & 69.60 & 85.71 & 83.70 & 96.44 \\
& aTLAS (N=40\%) & 64.26 & 96.93 & 26.70 & 72.30 & 85.94 & 88.06 & 97.25 \\
& AXIS (N=10\%) & 68.14 & 98.00 & 31.95 & 76.65 & 86.15 & 90.02 & 98.02 \\
& AXIS (N=20\%) & 70.85 & 98.33 & 29.70 & 79.10 & 86.49 & 93.61 & 98.54 \\
& AXIS (N=40\%) & 71.91 & 98.19 & 19.20 & 77.82 & 87.07 & 94.73 & 98.96 \\
\midrule
\multirow{6}{*}{15} & aTLAS (N=10\%) & 56.44 & 95.15 & 24.93 & 70.22 & 85.60 & 78.31 & 96.15 \\
& aTLAS (N=20\%) & 60.21 & 96.11 & 25.86 & 73.61 & 85.99 & 83.08 & 96.94 \\
& aTLAS (N=40\%) & 62.71 & 96.81 & 28.14 & 74.48 & 86.17 & 87.39 & 97.06 \\
& AXIS (N=10\%) & 67.82 & 97.78 & 31.05 & 77.25 & 86.18 & 91.00 & 98.20 \\
& AXIS (N=20\%) & 70.59 & 98.19 & 34.92 & 82.09 & 86.61 & 93.67 & 98.70 \\
& AXIS (N=40\%) & 71.38 & 98.26 & 39.15 & 83.67 & 87.11 & 94.76 & 98.89 \\
\midrule
\multirow{6}{*}{21} & aTLAS (N=10\%) & 56.44 & 95.07 & 25.62 & 71.23 & 85.72 & 78.02 & 95.98 \\
& aTLAS (N=20\%) & 60.37 & 96.26 & 26.37 & 72.09 & 85.91 & 83.45 & 96.94 \\
& aTLAS (N=40\%) & 63.24 & 96.96 & 26.25 & 75.09 & 86.29 & 88.38 & 97.58 \\
& AXIS (N=10\%) & 67.98 & 97.81 & 30.75 & 77.87 & 86.32 & 91.06 & 98.11 \\
& AXIS (N=20\%) & 70.64 & 98.22 & 34.50 & 82.31 & 86.57 & 93.46 & 98.64 \\
& AXIS (N=40\%) & 72.18 & 98.52 & 38.97 & 83.74 & 87.15 & 94.43 & 98.96 \\
\bottomrule
\end{tabular}
\end{adjustbox}
\label{tab:results_part2}
\end{table*}

\begin{table*}[!htbp]
\centering
\caption{Detailed results per target dataset for various numbers of source task vectors (STV). Part 3 of 3.}
\begin{adjustbox}{width=0.8\textwidth,center}
\begin{tabular}{@{}clccccccc@{}}
\toprule
\textbf{STV} & \textbf{Method} & \textbf{OxfordIIITPet} & \textbf{PascalVOC} & \textbf{RESISC45} & \textbf{STL10} & \textbf{SUN397} & \textbf{SVHN} & \textbf{UCF101} \\
\midrule
\multirow{6}{*}{1} & aTLAS (N=10\%) & 90.19 & 82.99 & 71.19 & 97.99 & 64.42 & 62.10 & 65.05 \\
& aTLAS (N=20\%) & 90.73 & 84.21 & 72.14 & 98.16 & 64.95 & 67.11 & 65.61 \\
& aTLAS (N=40\%) & 90.71 & 86.51 & 80.40 & 98.49 & 66.16 & 86.49 & 68.94 \\
& AXIS (N=10\%) & 89.92 & 85.77 & 87.51 & 97.65 & 66.80 & 86.63 & 71.00 \\
& AXIS (N=20\%) & 89.86 & 86.77 & 89.95 & 97.80 & 68.48 & 89.76 & 74.91 \\
& AXIS (N=40\%) & 90.24 & 86.53 & 91.84 & 97.08 & 70.05 & 91.35 & 77.98 \\
\midrule
\multirow{6}{*}{5} & aTLAS (N=10\%) & 90.62 & 85.49 & 74.56 & 97.91 & 64.85 & 83.23 & 65.95 \\
& aTLAS (N=20\%) & 91.31 & 86.16 & 77.16 & 98.35 & 65.43 & 84.09 & 67.94 \\
& aTLAS (N=40\%) & 91.99 & 86.72 & 80.43 & 98.34 & 66.29 & 86.34 & 68.86 \\
& AXIS (N=10\%) & 90.60 & 86.71 & 87.90 & 97.74 & 67.27 & 90.87 & 71.45 \\
& AXIS (N=20\%) & 90.19 & 87.09 & 90.41 & 97.65 & 68.69 & 92.18 & 76.37 \\
& AXIS (N=40\%) & 90.27 & 86.99 & 91.90 & 97.26 & 69.75 & 92.87 & 78.03 \\
\midrule
\multirow{6}{*}{10} & aTLAS (N=10\%) & 91.77 & 86.19 & 79.13 & 98.24 & 66.33 & 85.66 & 67.57 \\
& aTLAS (N=20\%) & 91.61 & 86.63 & 82.16 & 98.21 & 66.76 & 87.45 & 69.36 \\
& aTLAS (N=40\%) & 91.50 & 87.11 & 84.87 & 98.24 & 67.04 & 89.06 & 71.66 \\
& AXIS (N=10\%) & 90.11 & 86.50 & 88.38 & 97.73 & 67.28 & 87.92 & 73.17 \\
& AXIS (N=20\%) & 90.32 & 87.05 & 90.48 & 97.59 & 68.91 & 90.65 & 77.37 \\
& AXIS (N=40\%) & 89.53 & 86.75 & 92.75 & 96.86 & 70.41 & 92.61 & 78.09 \\
\midrule
\multirow{6}{*}{15} & aTLAS (N=10\%) & 91.63 & 86.87 & 78.79 & 98.50 & 66.43 & 85.62 & 68.68 \\
& aTLAS (N=20\%) & 92.78 & 87.39 & 82.40 & 98.70 & 67.48 & 87.66 & 70.90 \\
& aTLAS (N=40\%) & 92.18 & 87.62 & 84.97 & 98.53 & 67.82 & 89.14 & 72.51 \\
& AXIS (N=10\%) & 91.09 & 86.92 & 87.94 & 98.13 & 67.61 & 88.24 & 73.17 \\
& AXIS (N=20\%) & 91.03 & 87.64 & 90.54 & 97.89 & 68.70 & 89.97 & 75.87 \\
& AXIS (N=40\%) & 90.22 & 87.18 & 92.22 & 97.68 & 70.56 & 92.91 & 77.64 \\
\midrule
\multirow{6}{*}{21} & aTLAS (N=10\%) & 92.23 & 87.11 & 80.52 & 98.36 & 66.63 & 86.83 & 70.05 \\
& aTLAS (N=20\%) & 92.61 & 87.56 & 81.25 & 98.55 & 66.99 & 87.69 & 71.35 \\
& aTLAS (N=40\%) & 92.91 & 88.15 & 84.40 & 98.55 & 67.88 & 89.14 & 73.09 \\
& AXIS (N=10\%) & 91.25 & 87.25 & 88.25 & 98.05 & 67.62 & 88.56 & 74.28 \\
& AXIS (N=20\%) & 90.81 & 87.46 & 90.86 & 97.95 & 68.96 & 90.38 & 77.35 \\
& AXIS (N=40\%) & 90.71 & 86.97 & 91.97 & 97.30 & 70.29 & 92.64 & 79.96 \\
\bottomrule
\end{tabular}
\end{adjustbox}
\label{tab:results_part3}
\end{table*}

\begin{table*}[!htbp]
\centering
\caption{A detailed, per-dataset performance comparison of different SVD component aggregation strategies. The table reports the Top-1 accuracy (\%) for each target task, illustrating how performance evolves as the number of aggregated source task vectors (TV) increases. We compare our primary top components strategy against bottom components and arbitrary components as baselines to validate the robustness of our selection method across diverse data domains.}
\begin{adjustbox}{width=\textwidth}
\small
\begin{tabular}{@{}llccccccccccccccccccccc@{}}
\ TV & Strategy & \rotatebox{90}{CIFAR100} & \rotatebox{90}{CIFAR10} & \rotatebox{90}{CUB200} & \rotatebox{90}{Caltech101} & \rotatebox{90}{Caltech256} & \rotatebox{90}{Cars} & \rotatebox{90}{Country211} & \rotatebox{90}{DTD} & \rotatebox{90}{EuroSAT} & \rotatebox{90}{FGVCAircraft} & \rotatebox{90}{Flowers102} & \rotatebox{90}{Food101} & \rotatebox{90}{GTSRB} & \rotatebox{90}{MNIST} & \rotatebox{90}{OxfordIIITPet} & \rotatebox{90}{PascalVOC} & \rotatebox{90}{RESISC45} & \rotatebox{90}{STL10} & \rotatebox{90}{SUN397} & \rotatebox{90}{SVHN} & \rotatebox{90}{UCF101} \\
\cmidrule(r){1-2} \cmidrule(l){3-23}

1 & bottom components & 76.58 & 95.95 & 58.15 & 94.64 & 88.52 & 64.07 & 17.59 & 67.87 & 97.78 & 29.61 & 79.62 & 85.20 & 87.87 & 97.31 & 91.09 & 86.78 & 88.73 & 98.20 & 66.90 & 84.79 & 72.98 \\
  & top components & 77.00 & 95.85 & 57.61 & 93.89 & 88.44 & 63.54 & 17.70 & 67.02 & 97.30 & 29.70 & 77.49 & 85.81 & 89.57 & 97.36 & 89.92 & 85.77 & 87.51 & 97.65 & 66.80 & 86.63 & 71.00 \\
\midrule
2 & bottom components & 77.27 & 95.96 & 58.42 & 94.70 & 88.52 & 63.97 & 17.82 & 65.48 & 97.74 & 29.73 & 79.31 & 85.68 & 88.58 & 97.67 & 91.44 & 87.20 & 88.16 & 98.15 & 67.44 & 86.38 & 74.17 \\
  & arbitrary components & 77.14 & 95.97 & 57.99 & 94.64 & 87.90 & 63.56 & 17.77 & 65.37 & 97.70 & 29.52 & 77.67 & 85.69 & 88.79 & 97.55 & 91.52 & 87.17 & 87.14 & 98.08 & 66.64 & 85.68 & 72.64 \\
  & top components & 77.81 & 96.18 & 57.44 & 93.84 & 87.70 & 63.51 & 17.75 & 67.45 & 96.89 & 29.43 & 75.18 & 85.88 & 90.40 & 97.83 & 89.94 & 86.61 & 88.11 & 97.96 & 66.90 & 87.81 & 73.06 \\
\midrule
3 & bottom components & 77.51 & 95.75 & 58.78 & 95.10 & 88.32 & 64.63 & 17.75 & 66.17 & 97.67 & 29.85 & 80.52 & 85.71 & 88.27 & 97.63 & 91.14 & 86.89 & 88.00 & 98.06 & 67.20 & 85.68 & 73.80 \\
  & arbitrary components & 77.46 & 96.12 & 57.90 & 94.41 & 87.85 & 63.79 & 17.88 & 66.17 & 97.93 & 29.49 & 78.06 & 85.54 & 88.36 & 97.57 & 90.76 & 86.36 & 87.73 & 98.10 & 66.43 & 86.51 & 71.85 \\
  & top components & 77.37 & 96.08 & 57.59 & 93.84 & 87.99 & 64.21 & 17.84 & 66.17 & 97.37 & 30.03 & 77.07 & 85.79 & 89.72 & 97.69 & 90.02 & 86.28 & 87.48 & 97.71 & 67.60 & 87.28 & 72.85 \\
\midrule
4 & bottom components & 77.36 & 96.11 & 58.35 & 94.47 & 88.37 & 64.18 & 17.91 & 67.02 & 97.63 & 30.30 & 80.19 & 85.77 & 88.61 & 97.75 & 91.11 & 86.87 & 87.37 & 98.21 & 67.11 & 85.56 & 73.25 \\
  & arbitrary components & 77.17 & 96.13 & 58.32 & 94.70 & 87.99 & 63.96 & 17.81 & 66.22 & 97.11 & 30.39 & 79.35 & 85.56 & 87.78 & 97.69 & 91.01 & 86.75 & 87.19 & 98.21 & 67.06 & 86.28 & 73.09 \\
  & top components & 78.32 & 96.01 & 57.89 & 93.15 & 88.32 & 64.23 & 17.91 & 66.86 & 97.85 & 29.82 & 77.72 & 85.88 & 90.32 & 98.00 & 90.11 & 86.79 & 87.41 & 97.84 & 67.35 & 89.26 & 73.80 \\
\midrule
5 & bottom components & 77.62 & 95.83 & 57.70 & 94.59 & 88.48 & 64.08 & 17.78 & 65.90 & 97.41 & 29.49 & 78.86 & 85.74 & 88.95 & 97.50 & 90.73 & 86.73 & 88.02 & 98.31 & 66.86 & 84.98 & 72.75 \\
  & arbitrary components & 77.04 & 96.19 & 58.01 & 94.12 & 87.98 & 64.07 & 17.73 & 66.70 & 97.74 & 30.18 & 77.82 & 85.75 & 89.49 & 97.81 & 90.49 & 87.05 & 88.02 & 98.21 & 66.63 & 87.38 & 73.09 \\
  & top components & 77.51 & 96.50 & 58.41 & 93.61 & 88.01 & 63.95 & 18.17 & 65.69 & 97.41 & 30.48 & 77.22 & 86.05 & 90.74 & 97.78 & 90.60 & 86.71 & 87.90 & 97.74 & 67.27 & 90.87 & 71.45 \\
\midrule
6 & bottom components & 78.01 & 95.96 & 58.25 & 94.82 & 88.73 & 64.28 & 18.22 & 66.81 & 97.70 & 30.42 & 79.48 & 85.89 & 88.87 & 97.70 & 91.28 & 87.17 & 88.02 & 98.19 & 66.94 & 86.08 & 72.75 \\
  & arbitrary components & 77.54 & 96.17 & 57.99 & 94.53 & 88.08 & 64.21 & 17.82 & 65.32 & 97.89 & 30.45 & 78.48 & 85.77 & 88.43 & 97.50 & 90.84 & 86.69 & 87.92 & 98.01 & 67.09 & 87.58 & 71.42 \\
  & top components & 77.63 & 96.11 & 58.46 & 94.35 & 88.68 & 65.03 & 18.45 & 66.91 & 97.56 & 30.09 & 76.94 & 85.91 & 90.73 & 98.02 & 90.38 & 86.31 & 88.19 & 97.91 & 67.90 & 90.84 & 72.01 \\
\midrule
7 & bottom components & 78.04 & 95.93 & 58.44 & 94.87 & 88.52 & 64.51 & 17.77 & 65.37 & 97.59 & 30.03 & 78.37 & 85.94 & 88.57 & 97.39 & 91.28 & 87.07 & 88.02 & 98.08 & 66.88 & 86.09 & 73.43 \\
  & arbitrary components & 77.92 & 95.83 & 57.94 & 95.28 & 87.83 & 64.35 & 17.99 & 65.96 & 97.63 & 28.92 & 78.37 & 85.90 & 88.73 & 97.81 & 90.65 & 86.97 & 87.79 & 98.28 & 66.92 & 85.92 & 73.38 \\
  & top components & 78.12 & 96.20 & 57.99 & 94.41 & 88.22 & 64.72 & 17.82 & 67.61 & 97.81 & 30.51 & 77.54 & 86.15 & 91.43 & 98.36 & 89.97 & 86.53 & 88.22 & 97.80 & 67.43 & 90.80 & 71.95 \\
\midrule
8 & bottom components & 78.16 & 95.97 & 58.13 & 94.53 & 88.58 & 64.59 & 17.96 & 65.80 & 97.81 & 30.36 & 80.24 & 85.74 & 88.50 & 97.35 & 91.47 & 86.91 & 87.95 & 98.41 & 67.41 & 86.21 & 73.94 \\
  & arbitrary components & 77.56 & 96.07 & 57.34 & 93.95 & 88.01 & 63.87 & 18.00 & 66.44 & 97.67 & 30.21 & 78.63 & 85.69 & 88.73 & 97.61 & 91.41 & 86.73 & 87.62 & 98.00 & 67.24 & 86.44 & 74.60 \\
  & top components & 79.05 & 96.45 & 58.42 & 93.84 & 88.91 & 64.64 & 18.04 & 66.65 & 97.59 & 30.75 & 79.13 & 86.15 & 91.44 & 98.39 & 90.73 & 86.88 & 88.48 & 97.80 & 67.43 & 90.96 & 73.14 \\
\midrule
9 & bottom components & 78.29 & 96.12 & 58.30 & 93.95 & 88.39 & 64.51 & 17.92 & 66.65 & 97.74 & 30.90 & 80.00 & 85.79 & 88.16 & 97.70 & 91.50 & 87.15 & 87.90 & 98.26 & 67.14 & 86.12 & 73.80 \\
  & arbitrary components & 77.95 & 96.13 & 56.94 & 94.64 & 88.50 & 64.11 & 17.85 & 67.45 & 96.96 & 29.43 & 78.50 & 85.86 & 88.49 & 97.85 & 90.57 & 87.03 & 87.84 & 98.20 & 67.04 & 86.20 & 73.09 \\
  & top components & 79.42 & 96.83 & 58.47 & 95.74 & 88.40 & 64.57 & 18.32 & 67.50 & 97.59 & 30.93 & 78.08 & 86.17 & 92.24 & 98.50 & 90.32 & 86.87 & 88.46 & 97.59 & 67.11 & 91.00 & 72.46 \\
\midrule
10 & bottom components & 77.86 & 96.12 & 57.49 & 95.10 & 88.66 & 64.36 & 18.03 & 66.01 & 98.04 & 30.39 & 80.37 & 85.95 & 88.28 & 97.59 & 91.50 & 87.13 & 87.56 & 98.11 & 67.37 & 85.85 & 73.49 \\
   & arbitrary components & 77.27 & 96.08 & 57.99 & 94.64 & 88.32 & 63.98 & 17.91 & 65.37 & 97.44 & 30.54 & 79.51 & 85.96 & 88.23 & 97.83 & 90.98 & 87.06 & 87.71 & 98.11 & 67.31 & 85.98 & 72.69 \\
   & top components & 80.09 & 96.96 & 57.85 & 94.82 & 88.76 & 64.66 & 18.24 & 68.14 & 98.00 & 31.95 & 76.65 & 86.15 & 90.02 & 98.02 & 90.11 & 86.50 & 88.38 & 97.73 & 67.28 & 87.92 & 73.17 \\
\midrule
11 & bottom components & 77.84 & 95.81 & 57.80 & 93.84 & 88.75 & 64.48 & 18.06 & 67.34 & 97.81 & 30.78 & 80.00 & 85.95 & 87.68 & 97.80 & 91.47 & 87.00 & 88.03 & 98.29 & 67.39 & 86.34 & 73.78 \\
   & arbitrary components & 78.16 & 96.12 & 57.13 & 93.72 & 88.66 & 64.02 & 17.82 & 65.43 & 97.93 & 29.64 & 79.05 & 86.02 & 89.25 & 97.73 & 90.81 & 86.97 & 86.83 & 97.99 & 67.12 & 86.79 & 71.58 \\
   & top components & 80.14 & 96.97 & 59.11 & 95.39 & 88.48 & 64.51 & 18.39 & 66.97 & 97.96 & 30.45 & 78.55 & 86.15 & 90.39 & 98.17 & 90.98 & 87.37 & 87.92 & 97.78 & 67.47 & 88.99 & 74.25 \\
\midrule
12 & bottom components & 77.83 & 95.69 & 58.06 & 94.24 & 88.52 & 64.06 & 17.99 & 66.91 & 97.81 & 29.94 & 79.31 & 85.82 & 88.84 & 97.76 & 90.52 & 87.09 & 87.89 & 98.18 & 67.48 & 86.09 & 72.03 \\
   & arbitrary components & 78.18 & 96.29 & 57.58 & 93.89 & 88.44 & 64.22 & 17.75 & 67.66 & 97.85 & 30.93 & 78.18 & 86.01 & 87.66 & 97.72 & 90.92 & 86.89 & 87.43 & 98.08 & 67.06 & 86.60 & 73.62 \\
   & top components & 80.11 & 96.92 & 58.99 & 95.45 & 88.61 & 64.54 & 18.34 & 68.40 & 98.15 & 31.65 & 78.44 & 86.19 & 90.32 & 98.28 & 90.71 & 87.07 & 88.60 & 98.04 & 67.44 & 88.71 & 73.80 \\
\midrule
13 & bottom components & 77.82 & 95.89 & 58.53 & 94.18 & 88.86 & 64.12 & 18.20 & 67.55 & 97.93 & 29.70 & 78.73 & 85.90 & 87.64 & 97.57 & 91.50 & 87.07 & 87.84 & 98.31 & 67.42 & 86.25 & 72.72 \\
   & arbitrary components & 78.41 & 95.96 & 58.06 & 95.05 & 88.26 & 64.37 & 18.09 & 66.49 & 97.78 & 29.19 & 79.61 & 85.82 & 88.60 & 97.76 & 90.52 & 86.92 & 87.29 & 98.20 & 67.38 & 86.35 & 72.88 \\
   & top components & 80.05 & 96.83 & 58.85 & 94.70 & 88.81 & 65.10 & 18.38 & 67.39 & 98.19 & 30.84 & 76.81 & 86.17 & 90.28 & 98.18 & 91.36 & 87.26 & 88.54 & 98.05 & 67.55 & 88.84 & 73.62 \\
\midrule
14 & bottom components & 77.75 & 95.82 & 58.16 & 94.76 & 88.91 & 64.26 & 18.04 & 67.07 & 97.52 & 29.37 & 78.96 & 85.84 & 88.27 & 97.67 & 90.98 & 86.62 & 87.92 & 98.39 & 67.28 & 85.74 & 72.51 \\
   & arbitrary components & 78.21 & 96.00 & 57.63 & 93.95 & 88.32 & 64.57 & 18.09 & 67.29 & 97.59 & 30.69 & 79.92 & 86.02 & 88.14 & 97.88 & 91.01 & 87.08 & 87.78 & 98.08 & 66.98 & 86.53 & 73.38 \\
   & top components & 80.10 & 96.83 & 58.60 & 95.33 & 88.78 & 64.66 & 18.13 & 67.29 & 98.04 & 30.93 & 77.46 & 86.23 & 90.68 & 98.22 & 91.09 & 87.06 & 88.73 & 98.23 & 67.56 & 88.73 & 73.94 \\
\midrule
15 & bottom components & 78.01 & 95.79 & 58.01 & 93.95 & 88.96 & 64.23 & 17.93 & 67.13 & 97.19 & 29.43 & 78.63 & 85.92 & 88.47 & 97.64 & 90.65 & 87.01 & 87.78 & 98.45 & 67.27 & 85.55 & 72.51 \\
   & arbitrary components & 77.94 & 96.24 & 57.27 & 94.53 & 88.71 & 64.28 & 17.79 & 65.90 & 97.85 & 30.33 & 78.66 & 85.97 & 88.90 & 97.92 & 90.73 & 87.02 & 88.24 & 98.21 & 66.95 & 86.53 & 72.98 \\
   & top components & 80.14 & 96.85 & 58.68 & 94.64 & 88.65 & 65.43 & 18.31 & 67.82 & 97.78 & 31.05 & 77.25 & 86.18 & 91.00 & 98.20 & 91.09 & 86.92 & 87.94 & 98.13 & 67.61 & 88.24 & 73.17 \\
\midrule
16 & bottom components & 77.94 & 95.90 & 57.73 & 94.30 & 88.48 & 64.66 & 18.21 & 65.32 & 98.00 & 28.98 & 79.74 & 85.93 & 88.27 & 97.56 & 91.06 & 87.13 & 87.65 & 98.28 & 67.39 & 85.97 & 72.32 \\
   & arbitrary components & 78.40 & 96.26 & 57.70 & 94.70 & 88.29 & 64.59 & 18.22 & 67.98 & 97.78 & 30.63 & 80.09 & 86.06 & 88.70 & 97.85 & 91.11 & 86.76 & 88.16 & 98.31 & 67.05 & 87.45 & 72.72 \\
   & top components & 80.13 & 96.78 & 59.06 & 94.47 & 88.88 & 65.30 & 18.28 & 68.09 & 97.74 & 30.63 & 78.37 & 86.23 & 90.58 & 98.03 & 90.79 & 87.16 & 87.76 & 97.91 & 67.62 & 88.64 & 74.31 \\
\midrule
17 & bottom components & 77.64 & 95.87 & 57.85 & 94.99 & 88.30 & 64.40 & 17.89 & 66.49 & 97.70 & 30.45 & 80.47 & 85.99 & 88.18 & 97.78 & 90.73 & 87.35 & 88.03 & 98.35 & 67.60 & 85.91 & 72.64 \\
   & arbitrary components & 78.15 & 96.27 & 57.90 & 94.47 & 88.14 & 64.61 & 17.97 & 65.74 & 97.59 & 30.06 & 79.35 & 86.03 & 88.47 & 97.70 & 91.25 & 87.23 & 87.78 & 98.21 & 66.98 & 86.42 & 73.20 \\
   & top components & 80.17 & 96.91 & 58.77 & 95.45 & 88.89 & 65.03 & 18.27 & 66.76 & 97.89 & 30.12 & 78.96 & 86.23 & 90.82 & 98.10 & 91.20 & 86.91 & 87.75 & 97.95 & 67.67 & 88.38 & 74.31 \\
\midrule
18 & bottom components & 77.55 & 96.01 & 58.47 & 94.18 & 88.91 & 64.71 & 17.88 & 66.81 & 98.04 & 29.37 & 80.89 & 85.93 & 89.04 & 97.51 & 90.65 & 87.09 & 87.68 & 98.39 & 67.50 & 86.62 & 72.22 \\
   & arbitrary components & 78.34 & 96.14 & 57.85 & 94.07 & 88.37 & 65.10 & 18.05 & 66.76 & 97.44 & 29.04 & 80.68 & 86.10 & 88.92 & 97.77 & 91.41 & 87.23 & 87.27 & 98.36 & 67.20 & 86.65 & 73.22 \\
   & top components & 80.00 & 96.86 & 58.70 & 94.87 & 89.10 & 65.10 & 18.22 & 68.78 & 97.74 & 31.95 & 77.74 & 86.17 & 90.67 & 98.19 & 91.20 & 86.90 & 88.37 & 97.89 & 67.58 & 88.29 & 75.05 \\
\midrule
19 & bottom components & 77.76 & 95.92 & 58.56 & 94.18 & 88.47 & 64.43 & 17.91 & 66.44 & 97.89 & 30.57 & 79.85 & 85.85 & 88.85 & 97.68 & 90.43 & 87.25 & 88.03 & 98.44 & 67.45 & 85.51 & 72.32 \\
   & arbitrary components & 77.71 & 96.37 & 58.32 & 94.82 & 88.89 & 65.00 & 17.95 & 66.28 & 97.89 & 29.79 & 79.28 & 85.86 & 89.68 & 97.97 & 90.68 & 86.99 & 87.52 & 98.20 & 67.05 & 86.93 & 72.91 \\
   & top components & 79.98 & 96.89 & 59.15 & 95.22 & 88.97 & 64.86 & 18.46 & 67.87 & 98.07 & 32.40 & 77.82 & 86.25 & 90.89 & 98.07 & 90.98 & 87.19 & 88.33 & 98.05 & 67.79 & 88.18 & 74.73 \\
\midrule
20 & bottom components & 77.62 & 96.04 & 58.61 & 94.12 & 88.45 & 64.81 & 17.94 & 66.54 & 98.00 & 29.91 & 79.67 & 85.88 & 88.65 & 97.84 & 90.84 & 87.25 & 88.02 & 98.44 & 67.49 & 86.67 & 72.96 \\
   & arbitrary components & 77.72 & 95.99 & 57.80 & 94.47 & 88.29 & 65.08 & 17.86 & 66.91 & 97.56 & 29.88 & 79.46 & 86.23 & 88.73 & 97.88 & 91.09 & 86.79 & 87.38 & 98.24 & 66.89 & 86.42 & 72.48 \\
   & top components & 80.07 & 96.93 & 58.94 & 95.10 & 88.88 & 64.83 & 18.44 & 68.94 & 97.70 & 30.00 & 78.03 & 86.35 & 90.69 & 98.11 & 91.14 & 87.37 & 88.29 & 97.99 & 67.65 & 88.30 & 74.86 \\
\midrule
21 & bottom components & 77.57 & 95.85 & 58.63 & 94.07 & 88.65 & 64.74 & 18.19 & 65.37 & 97.48 & 30.30 & 79.74 & 85.97 & 87.06 & 97.82 & 90.84 & 87.23 & 88.46 & 98.46 & 67.52 & 85.48 & 73.28 \\
   & arbitrary components & 78.05 & 96.16 & 57.61 & 94.99 & 87.98 & 64.82 & 17.98 & 66.49 & 97.22 & 29.64 & 77.65 & 86.08 & 88.52 & 97.55 & 90.60 & 86.96 & 87.38 & 98.28 & 67.11 & 87.73 & 73.88 \\
   & top components & 80.11 & 96.93 & 58.46 & 94.99 & 88.76 & 65.09 & 18.48 & 67.98 & 97.81 & 30.75 & 77.87 & 86.32 & 91.06 & 98.11 & 91.25 & 87.25 & 88.25 & 98.05 & 67.62 & 88.56 & 74.28 \\

\bottomrule
\end{tabular}
\end{adjustbox}
\label{tab:appendix_full_results_comparison}
\end{table*}


\end{document}